\documentclass[acmlarge]{acmart}

\usepackage{multirow}
\usepackage{rotating}
\usepackage{makecell}
\usepackage{longtable}
\AtBeginDocument{%
  }

\setcopyright{acmlicensed}
\copyrightyear{2025}
\acmYear{2025}
\acmDOI{XXXXXXX.XXXXXXX}
\acmJournal{HEALTH}
\acmVolume{1}
\acmNumber{1}
\acmArticle{1}
\acmMonth{1}

\begin{document}

\title{Explainable Graph Neural Networks: Understanding Brain Connectivity and Biomarkers in Dementia}



\author{Niharika Tewari}
\orcid{0000-0002-7690-9814}
\affiliation{%
    \department{School of Computing Technologies}
    \institution{RMIT University}
    \city{Melbourne}
    \state{VIC}
    \country{Australia}
}
\affiliation{%
    \department{Department of Biological Sciences and Department of Computer Science \& Information Systems}
    \institution{Birla Institute of Technology and Science, Pilani, K K Birla Goa Campus}
    \city{Zuarinagar, Sancoale}
    \state{Goa, 403726}
    \country{India}
}
\email{s4044597@student.rmit.edu.au}

\author{Nguyen Linh Dan Le}
\orcid{0009-0008-5888-0932}
\affiliation{%
    \department{School of Computing Technologies}
    \institution{RMIT University}
    \city{Melbourne}
    \state{VIC}
    \country{Australia}}
\email{s4135772@student.rmit.edu.au}

\author{Mujie Liu}
\orcid{0009-0002-0879-7168}
\affiliation{%
\department{Institute of Innovation, Science and Sustainability}
    \institution{Federation University Australia}
    \city{Ballarat}
    \state{VIC}
    \country{Australia}
}
\email{mujie.liu@ieee.org}

\author{Jing Ren}
\orcid{0000-0003-0169-1491}
\affiliation{%
\department{School of Computing Technologies}
    \institution{RMIT University}
    \city{Melbourne}
    \state{VIC}
    \country{Australia}
}
\email{jing.ren@ieee.org}

\author{Ziqi Xu}
\orcid{0000-0003-1748-5801}
\authornote{Corresponding Author}
\affiliation{%
\department{School of Computing Technologies}
    \institution{RMIT University}
    \city{Melbourne}
    \state{VIC}
    \country{Australia}
}
\email{ziqi.xu@rmit.edu.au}

\author{Tabinda Sarwar}
\orcid{0000-0001-7313-5350}
\affiliation{%
\department{School of Computing Technologies}
    \institution{RMIT University}
    \city{Melbourne}
    \state{VIC}
    \country{Australia}
}
\email{tabinda.sarwar@rmit.edu.au}

\author{Veeky Baths}
\orcid{0000-0001-9980-5738}
\affiliation{%
    \department{Department of Biological Sciences and Department of Computer Science \& Information Systems}
    \institution{Birla Institute of Technology and Science, Pilani, K K Birla Goa Campus}
    \city{Zuarinagar, Sancoale}
    \state{Goa, 403726}
    \country{India}
}
\email{veeky@goa.bits-pilani.ac.in}

\author{Feng Xia}
\orcid{0000-0002-8324-1859}
\affiliation{%
    \department{School of Computing Technologies}
    \institution{RMIT University}
    \city{Melbourne}
    \state{VIC}
    \country{Australia}
}
\email{f.xia@ieee.org}


\begin{abstract}
Dementia is a progressive neurodegenerative disorder with multiple etiologies, including Alzheimer’s disease, Parkinson’s disease, frontotemporal dementia, and vascular dementia. Its clinical and biological heterogeneity makes diagnosis and subtype differentiation highly challenging. Graph Neural Networks (GNNs) have recently shown strong potential in modeling brain connectivity, but their limited robustness, data scarcity, and lack of interpretability constrain clinical adoption. Explainable Graph Neural Networks (XGNNs) have emerged to address these barriers by combining graph-based learning with interpretability, enabling the identification of disease-relevant biomarkers, analysis of brain network disruptions, and provision of transparent insights for clinicians. This paper presents the first comprehensive review dedicated to XGNNs in dementia research. We examine their applications across Alzheimer’s disease, Parkinson’s disease, mild cognitive impairment, and multi-disease diagnosis. A taxonomy of explainability methods tailored for dementia-related tasks is introduced, alongside comparisons of existing models in clinical scenarios. We also highlight challenges such as limited generalizability, underexplored domains, and the integration of Large Language Models (LLMs) for early detection. By outlining both progress and open problems, this review aims to guide future work toward trustworthy, clinically meaningful, and scalable use of XGNNs in dementia research.
\end{abstract}

\ccsdesc[500]{Computing methodologies~Neural networks}
\ccsdesc[500]{Applied computing~Health informatics}
\ccsdesc[500]{Applied computing~Computational biology}
\keywords{Explainable Graph Neural Networks, Dementia, Health informatics, Brain connectivity}


\maketitle

\section{Introduction}

Dementia, encompassing a range of conditions such as Alzheimer’s disease, vascular dementia, frontotemporal dementia, and Parkinson’s disease dementia, has become a pressing global health challenge. It is estimated that around fifty million people worldwide are currently living with dementia of various etiologies, a figure that is projected to triple by 2050, with the highest mortality rates occurring in low- and middle-income countries \cite{bib1}. In the absence of a cure, the rising prevalence of dementia places an increasing strain on healthcare systems. Early diagnosis is vital, as it enables timely interventions such as lifestyle modifications, pharmacological treatment, and cognitive therapy measures that can reduce the global incidence of dementia by as much as 40\% \cite{bib2}. However, achieving accurate and timely diagnosis remains challenging due to the complexity and heterogeneity of clinical presentations. In this context, the integration of artificial intelligence (AI) into dementia research has significantly enhanced early detection and treatment efforts by enabling the analysis of large-scale data, predicting disease progression, supporting clinical decision-making, and reinforcing healthcare infrastructure \cite{bib3}.

Assessing cognitive function in patients with dementia is inherently complex due to the interplay between primary symptoms that directly result from brain damage and secondary manifestations reflecting the individual’s adaptation to cognitive impairment. Dementia follows a biologically progressive trajectory, initially affecting specific brain regions and giving rise to memory deficits, diminished cognitive abilities, mood instability, and visual-perceptual difficulties. The prodromal phase, also referred to as mild cognitive impairment (MCI) or pre-dementia, can often be prolonged, during which symptoms gradually intensify, including language impairments and perceptual deficits. In the final stage, dementia exerts a profound impact on daily functioning, with symptoms across different dementia types converging; this stage typically lasts one to two years \cite{bib4}.

More than fifty medical conditions have been identified as potential causes of dementia, among which Alzheimer’s disease (AD) remains the most prevalent, accounting for 60–80\% of cases. Frontotemporal dementia (FTD) affects approximately 60\% of younger individuals aged 45 to 60, but only around 3\% of older adults \cite{bib5}. Lewy body dementia (LwD) constitutes roughly 5\% of cases, predominantly among older populations. In addition, about 3.6\% of patients are diagnosed with Parkinson’s disease (PD), while Parkinson’s disease dementia (PDD) develops in approximately 24\% of those with PD \cite{bib6}. Mixed or multi-etiology dementia, which refers to the coexistence of multiple pathological features, is increasingly recognized in clinical practice. Notably, over 50\% of individuals diagnosed with AD also exhibit signs of FTD, vascular dementia (VaD), or other overlapping pathologies \cite{bib7, bib8}.

For diagnostic purposes, computational methods such as machine learning (ML) and deep learning (DL) have been employed to analyze disease patterns using medical imaging and clinical records. While these approaches have advanced dementia research, they often fall short in capturing the complex patterns of brain connectivity. Beyond the mere detection of abnormalities, it is crucial to understand how dementia affects specific brain regions and disrupts their interconnections. To address this limitation, Graph Neural Networks (GNNs) provide a more specialized solution by leveraging the inherent graph structure of brain networks. This allows them to model inter-regional relationships, structural disruptions, and changes in connectivity more effectively \cite{bib10,yu2024long}. Unlike convolutional neural networks (CNNs), which operate on connectome adjacency matrices but often lack specificity, GNNs are capable of learning meaningful node representations and extracting disease-relevant topological features \cite{bib11}. By representing brain data as graphs, GNNs enable a more precise characterization of brain connectivity, which is particularly valuable in dementia research for identifying subtypes, analyzing dynamic connectivity, and discovering potential biomarkers.

\begin{figure}[t]
    \centering
    \includegraphics[width=0.9\textwidth]{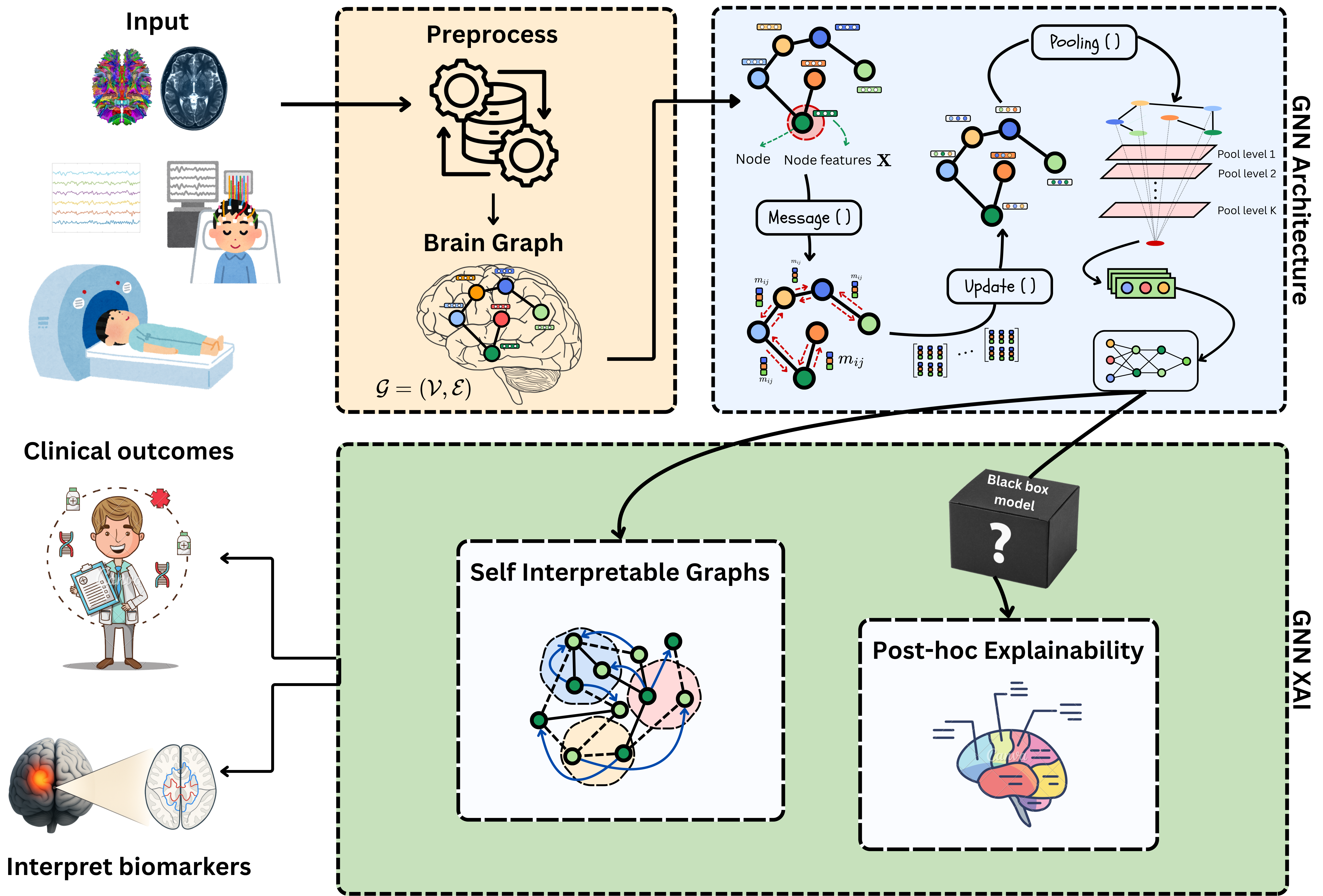}
    \caption{Visual representation of the XGNN framework used in studies.}
    \label{fig1}
    \Description{The figure depicts the architecture of the XGNN framework. 
    Brain connectivity graphs are processed through graph neural network layers, 
    then an explainability module highlights relevant regions of interest 
    while predicting clinical outcomes.}
\end{figure}

The application of GNNs has recently attracted considerable attention due to their strong predictive capabilities, particularly when applied to graph-structured biological data. While promising results have been achieved on smaller datasets, several challenges limit their clinical applicability. These include limited robustness and generalizability, a scarcity of labeled datasets, and potential biases arising from training on cohort-specific data. Such limitations undermine performance across diverse populations. Furthermore, dementia encompasses distinct etiologies such as PD and FTD, each with unique pathological and symptomatic profiles. The heterogeneous nature of dementia further complicates cross-cohort generalization \cite{bib13}.

In addition, debate persists regarding the effectiveness of graph-based representations in fully capturing the complexities of brain activity and in identifying appropriate methods for modeling brain connectivity. Trust and fairness are also critical considerations in healthcare, where transparent decision-making is essential to foster clinician confidence \cite{bib14, ryan2025fairness, trustworthiness}. A lack of interpretability can undermine this trust, which is vital for informed medical decisions \cite{bib15, bib16}. To address these concerns, Explainable Artificial Intelligence (XAI) has emerged as a promising approach to enhance model transparency and clarify the decision-making processes of complex systems. Despite these challenges, recent research underscores the potential of Explainable Graph Neural Networks (XGNNs) in improving dementia classification and prediction.

\subsection{Related Surveys}

Numerous studies have explored the application of machine learning and deep learning techniques in medical imaging \cite{patricio2023explainable, bhati2024survey, saranya2023systematic, medimaging, van2022explainable, muhammad2024unveiling}, disease diagnosis \cite{bib113trade}, and dementia research \cite{bib17, bib3, wahul2025multimodal}, offering broad insights into model architectures and clinical applications. With the emergence of explainability in GNNs \cite{yuan2022explainability, bib117xai}, there has been growing adoption of XGNNs in diagnosing brain disorders \cite{bib157}, investigating cognitive dysfunctions \cite{bib19}, and studying neurodegenerative diseases such as AD \cite{bib18, arya2023systematic} and PD \cite{jiang2024power}. These developments highlight the increasing need for a deeper understanding of dementia through the lens of XGNNs. As illustrated in Figure~\ref{fig1}, the XGNN framework typically combines graph-based learning with explainability modules to provide interpretable insights into brain connectivity and disease mechanisms.

Despite these advances, existing reviews often lack a focused discussion on the relevance and application of XGNN models to dementia research, which limits a clear understanding of the significance of this emerging intersection. To address this gap, our study presents a comprehensive survey of XGNN approaches in the context of dementia-related disorders, with a particular focus on both post-hoc and inherently interpretable models.

\subsection{Contributions}

This study aims to provide readers with a clear understanding of how XGNNs are applied in dementia research. We present a comprehensive overview of current applications, associated challenges, and future research directions, with the aim of promoting the broader adoption of XGNNs in the study of dementia. To the best of our knowledge, this is the first review dedicated to the use of XGNNs for understanding brain connectivity and identifying biomarkers in dementia. The key contributions of this work are summarized as follows:

\begin{itemize}
\item An exploration of the potential applications of XGNNs in dementia research. This includes individual disorders such as AD, PD, and MCI, as well as multi-disease diagnosis involving combinations of these and related conditions.

\item A comprehensive review and categorization of recent developments in XGNN algorithms for dementia research. We introduce a taxonomy that classifies explainability techniques according to scope, methodological type, domain applicability, and output form, thereby providing a structured framework for their use across both single- and multi-disease contexts.

\item A comparative analysis of existing XGNN models in various clinical scenarios, highlighting their diagnostic performance across distinct dementia subtypes and complex multi-disease cases.

\item A discussion of current challenges and limitations in the application of XGNNs to dementia detection, such as underexplored domains, integration with Large Language Models (LLMs) for early detection, and strategies for enhancing diagnostic accuracy. These discussions help identify promising directions for future research.
\end{itemize}

The remainder of this paper is organized as follows. Section \ref{sec:section2} presents the preliminaries, laying the foundation for understanding the motivation behind this review. Section \ref{sec:section3} introduces a subtype taxonomy of dementia and the review motivation, offering a classification of common dementia-related disorders and highlighting the need for explainable methods. Section \ref{sec:section4} presents a taxonomy of XGNNs. The subsequent Sections~\ref{sec:section5}, \ref{sec:section6}, \ref{sec:section7}, and \ref{sec:section8} explore how XGNNs are applied to specific conditions, including mild cognitive impairment, Alzheimer’s disease, Parkinson’s disease, and multi-disease diagnosis. Section \ref{sec:section9} reviews the public datasets used in studies, which form the empirical basis for model development and evaluation. Section \ref{sec:section10} concludes with a discussion on open challenges and outlook, summarizing limitations and future directions. Finally, Section~\ref{sec:section11} offers concluding reflections.

\section{Preliminaries}
\label{sec:section2}

\subsection{Multimodal Data}

Accurate dementia diagnosis and subtype differentiation rely on a range of medical imaging modalities, each of which captures complementary aspects of neural structure and function:

\begin{itemize}
    \item Structural Magnetic Resonance Imaging (sMRI): Provides high-resolution anatomical images, allowing for the quantification of grey matter volume, cortical thickness, and white matter integrity. Structural atrophy in regions such as the hippocampus and prefrontal cortex serves as a key biomarker in conditions such as Alzheimer’s disease (AD), frontotemporal dementia (FTD), and others~\cite{bib22s}.
    \item Functional Magnetic Resonance Imaging (fMRI): Measures spontaneous low-frequency BOLD (Blood Oxygen Level Dependent) signal fluctuations to assess functional connectivity (FC) between brain regions. Altered FC patterns have been widely observed in MCI, AD, and PDD~\cite{bib67bf}.
    \item Diffusion Tensor Imaging (DTI): Captures microstructural changes in white matter by modeling water diffusion, aiding the detection of disrupted white matter tracts and subcortical damage, particularly in VaD and early-stage PD~\cite{brueggen2017european}.
    \item Electroencephalography (EEG): Offers high temporal resolution and reveals abnormal spectral patterns and brain rhythms associated with dementia. It is especially valuable in early-stage diagnosis and in low-resource clinical settings~\cite{jeong2004eeg}.
    \item Positron Emission Tomography (PET): Includes amyloid-PET and FDG-PET, enabling direct visualization of metabolic activity and pathological protein deposition (e.g., A$\beta$, tau). PET provides critical insights into underlying AD pathology~\cite{chetelat2020amyloid}.
\end{itemize}

These imaging modalities are often employed individually or integrated through multimodal fusion to enhance diagnostic accuracy and better characterize the complex neurodegenerative processes underlying dementia. A clear understanding of these modalities also provides the foundation for constructing brain graphs and developing explainable machine learning models for dementia research.

\subsection{Brain Parcellation}

To analyze regional brain properties and construct connectomes, the brain is typically parcellated into anatomically or functionally defined regions of interest (ROIs) using standardized brain atlases. Commonly adopted parcellation schemes include:

\begin{itemize}
    \item \textbf{Automated Anatomical Labeling (AAL) atlas:} A widely used structural atlas that partitions the brain into 90 regions (or 116 including the cerebellum) based on anatomical landmarks.
    
    \item \textbf{Desikan–Killiany atlas:} Derived from gyral morphology, this atlas segments the cortex into 68 regions (34 per hemisphere) and is frequently employed in FreeSurfer-based cortical thickness analysis.
    
    \item \textbf{Schaefer atlas:} A functional parcellation generated from resting-state fMRI data using gradient-weighted Markov Random Field modeling. It supports flexible resolution ranging from 100 to 1000 ROIs, aligned hierarchically with functional networks.
    
    \item \textbf{Destrieux atlas:} A high-resolution anatomical atlas based on detailed gyral and sulcal structures. It offers finer granularity than the Desikan–Killiany atlas and is particularly useful in surface-based morphometry.
    
    \item \textbf{Harvard–Oxford atlas:} A probabilistic anatomical atlas encompassing both cortical and subcortical regions. It is commonly applied in volumetric and voxel-wise analyses.
    
    \item \textbf{SPHARM-PDM atlas:} A surface-based atlas that captures shape features of subcortical structures using spherical harmonics. It is particularly valuable for morphological graph construction in shape analysis.
\end{itemize}

By defining consistent ROIs, these atlases provide standardized frameworks for cross-subject comparison and enable multimodal integration, forming the basis for brain graph construction in dementia research~\cite{parisot2018disease, arslan2018human,peng2025joint}.

\subsection{Graph Construction}

Brain graphs (or brain networks) provide a compact mathematical abstraction of brain connectivity, representing ROIs as nodes and their interrelationships as edges. Formally, a brain graph is denoted as $G = (V, E, \mathbf{A})$, where $V = \{v_1, v_2, \dots, v_N\}$ represents a set of $N$ brain regions (nodes), $E$ denotes the set of edges, and $\mathbf{A} \in \mathbb{R}^{N \times N}$ is the weighted adjacency matrix that encodes the strength or similarity of connectivity between pairs of ROIs. The construction of $\mathbf{A}$ depends on the imaging modality and the chosen connectivity estimation method.

In practice, multiple types of brain graphs can be derived from different neuroimaging modalities:

\begin{itemize}
    \item \textbf{Functional brain graphs} are typically constructed from resting-state or task-based fMRI and EEG data. Here, nodes correspond to predefined ROIs, and edges reflect statistical dependencies between the neural activity time series of each ROI pair. For a given subject, let $\mathbf{x}_i \in \mathbb{R}^T$ denote the time series of ROI $v_i$ over $T$ time points. The functional connectivity (FC) between $v_i$ and $v_j$ is commonly estimated using similarity measures. A widely used metric is the Pearson correlation coefficient:
    \begin{equation}
        a_{ij}^{\text{func}} = \frac{\text{Cov}(\mathbf{x}_i, \mathbf{x}_j)}{\sigma(\mathbf{x}_i) \cdot \sigma(\mathbf{x}_j)},
    \end{equation}
    where $a_{ij}^{\text{func}}$ is the edge weight in the functional adjacency matrix $\mathbf{A}^{\text{func}}$, $\text{Cov}(\cdot,\cdot)$ denotes covariance, and $\sigma(\cdot)$ is the standard deviation. Although Pearson correlation is favored for its simplicity and interpretability, alternative FC measures—such as mutual information, spectral coherence, and phase-locking value—may be employed depending on the modality’s temporal resolution~\cite{smith2011network, vecchio2014human}.

    \item \textbf{Structural brain graphs} are derived from structural MRI or diffusion-based imaging techniques such as DTI. In this context, edges represent physical or anatomical connections. For DTI-based graphs, tractography is used to estimate white matter tracts between ROIs, with edge weights defined by metrics such as the number of streamlines, mean fractional anisotropy (FA), or average fiber length. Alternatively, morphometric similarity networks can be constructed by computing inter-regional correlations of morphological features (e.g., cortical thickness, surface area, or volume) across subjects, producing a structural adjacency matrix $\mathbf{A}^{\text{struct}}$ that reflects structural covariance~\cite{lori2002diffusion, hagmann2008mapping}.

    \item \textbf{Morphological brain graphs} capture shape- and geometry-based relationships, typically using structural MRI alongside methods such as SPHARM (spherical harmonics) or surface-based analysis. In these graphs, nodes still correspond to anatomical regions, while edges encode similarities in regional morphology, such as curvature profiles, sulcal depth, or hippocampal shape descriptors. These features are particularly useful for detecting disease-related structural deformations, with the resulting adjacency matrix $\mathbf{A}^{\text{morph}}$ representing inter-regional morphological correlations or proximity~\cite{dale1999cortical, mahjoub2018brain}.
\end{itemize}

Overall, brain graphs provide a unified yet flexible framework for modeling different dimensions of brain organization. In dementia research, they serve as a powerful abstraction for exploring connectivity disruptions, identifying disease biomarkers, and modeling disease progression, thereby providing the foundation for graph-based machine learning approaches in dementia research.

\section{Subtype Taxonomy of Dementia}
\label{sec:section3}

Dementia comprises a group of neurodegenerative disorders characterized by the progressive deterioration of cognitive function, often accompanied by impairments in memory, language, attention, executive function, visuospatial abilities, and behavior. Within this spectrum, several clinically distinct subtypes have been widely recognized, each exhibiting unique symptom profiles and patterns of brain involvement. In this review, we focus on five representative subtypes: MCI, AD, VaD, FTD, and PDD.

Although all five subtypes are introduced to provide a comprehensive clinical and neuropathological overview, the model-focused analysis in subsequent sections primarily centers on MCI, AD, and PD. This emphasis reflects the larger and more targeted body of XGNN-related literature for these conditions. In contrast, FTD and VaD remain underrepresented in the current landscape. Accordingly, we discuss studies of these subtypes within other sections rather than dedicating separate subsections; for instance, research on subcortical ischemic vascular dementia (SIVD) is presented under non-amnestic MCI, while comparative analyses of AD and FTD appear in the multi-disease diagnosis section.

This structured approach maintains clarity while acknowledging the growing relevance of FTD and VaD in GNN-based dementia research. As the field evolves, future work is expected to address these underexplored subtypes more directly, thereby enriching the scope of graph-based methodologies in dementia studies.

\subsection{Mild Cognitive Impairment} 

MCI is a clinical condition characterized by cognitive decline greater than expected for an individual's age and education level, yet not severe enough to significantly impair daily functioning. It is widely regarded as a transitional stage between normal cognitive aging and dementia, particularly AD in its amnestic form~\cite{bib95mci, bib96mci}. From an etiological perspective, MCI may arise from various mechanisms, including early neurodegenerative changes (e.g., amyloid deposition, tau pathology), cerebrovascular abnormalities, metabolic dysfunction, and psychiatric conditions such as depression or anxiety~\cite{jack2018nia}. Genetic predispositions, such as the presence of the \textit{APOE}~$\varepsilon4$ allele, have also been linked to an elevated risk of progression from MCI to AD~\cite{risacher2015apoe}. Clinically, MCI is categorized into amnestic and non-amnestic subtypes, depending on whether memory impairment is the predominant feature. Amnestic MCI (aMCI), which primarily affects memory, is more likely to progress to AD, whereas non-amnestic MCI (naMCI) involves deficits in language, attention, or executive function and may precede other subtypes such as FTD, PDD, or VaD~\cite{dugger2015neuropathological}. Longitudinal studies suggest that individuals with MCI convert to dementia at an annual rate of 10--15\%, although some remain stable or revert to normal cognition~\cite{langa2014diagnosis}. Diagnosis typically relies on standardized cognitive screening tools such as the Montreal Cognitive Assessment (MoCA) or the Mini-Mental State Examination (MMSE), supplemented by neuropsychological assessments, with advanced neuroimaging techniques (MRI, PET) increasingly used for subtype differentiation and early risk identification. 

Neuroanatomically, aMCI is associated with structural atrophy and metabolic dysfunction in memory-related regions including the hippocampus, entorhinal cortex, parahippocampal gyrus, posterior cingulate cortex (PCC), precuneus, and broader medial temporal lobe (MTL)~\cite{jack2010hypothetical, albert2011diagnosis}. These regions overlap with the default mode network (DMN), whose early disruption is considered a hallmark of preclinical AD. In contrast, naMCI is characterized by impairments in executive, language, or visuospatial domains, with alterations commonly observed in the dorsolateral prefrontal cortex (DLPFC), inferior parietal lobule, superior temporal gyrus, and anterior cingulate cortex. In cases linked to small vessel disease or mixed pathology, subcortical structures such as the thalamus, caudate, and related white matter tracts may also be affected~\cite{tampi2015mild}. These distinct anatomical patterns support differentiation among dementia subtypes and may guide the development of subtype-specific interventions.

\subsection{Alzheimer’s Disease}  

AD is a progressive neurodegenerative disorder and the most prevalent cause of dementia, accounting for 60--80\% of cases globally. It is clinically characterized by a gradual decline across multiple cognitive domains, beginning with episodic memory and followed by impairments in language, orientation, executive function, and personality~\cite{bib83}. Unlike MCI, AD represents a fully developed clinical syndrome with substantial impairment in daily functioning. From a pathophysiological perspective, AD is multifactorial, involving a cascade of molecular and cellular events. Hallmark pathological features include the extracellular accumulation of amyloid-beta (A$\beta$) plaques, intracellular neurofibrillary tangles composed of hyperphosphorylated tau protein, synaptic loss, and widespread neuronal death~\cite{selkoe1996amyloid, jack2013tracking}. These processes disrupt network-level brain communication, particularly within memory and self-referential systems. Clinical diagnosis typically combines neuropsychological testing, MRI-based volumetrics, and amyloid/tau PET imaging to identify early-stage AD and differentiate it from non-AD dementias~\cite{jack2018nia}. 

Neuroimaging and postmortem studies consistently reveal a characteristic pattern of brain atrophy and functional disconnection in AD. Early degeneration occurs in the hippocampus and medial temporal lobe, resulting in profound memory deficits. As the disease progresses, structural and metabolic alterations extend to the posterior cingulate cortex (PCC) and precuneus, which are key hubs within the DMN and support autobiographical memory, introspection, and spatial navigation—functions frequently impaired in AD~\cite{sestieri2011episodic, greicius2004default}. AD also affects the temporoparietal cortex, which underpins semantic knowledge and language comprehension, distinguishing it from subtypes such as FTD that typically spare this region in early stages. Moreover, disruptions in limbic-associated regions (LIM), including the amygdala and orbitofrontal cortex, contribute to behavioral symptoms such as anxiety, apathy, and emotional dysregulation~\cite{balthazar2014neuropsychiatric, seeley2009neurodegenerative}. Overall, the selective vulnerability of the hippocampal–DMN–temporoparietal axis provides a reliable anatomical and functional signature that supports early diagnosis, disease staging, and the development of targeted interventions.

\subsection{Vascular Dementia}  

VaD is the second most common form of dementia after AD and results from cerebrovascular pathology that disrupts blood flow to the brain, leading to neuronal injury and cognitive decline. Unlike neurodegenerative dementias, VaD arises from ischemic or hemorrhagic events that cause direct tissue damage or network disconnection, making it a structurally driven rather than proteinopathy-driven disorder~\cite{bib84, kalaria2010vascular}. The etiology of VaD is highly heterogeneous, encompassing large vessel strokes, lacunar infarcts, strategic infarcts, microbleeds, cerebral small vessel disease, and chronic hypoperfusion. Risk factors include hypertension, diabetes, hyperlipidemia, atrial fibrillation, and smoking—conditions that contribute to both macrovascular and microvascular injury~\cite{bib87v, van2018vascular}. Clinically, VaD does not follow a uniform progression pattern: cognitive decline often occurs in a stepwise or fluctuating manner, especially in multi-infarct types, whereas subcortical ischemic VaD may present more gradually. Patients frequently exhibit impairments in attention, processing speed, executive function, and gait, which are distinct from the early memory loss typically seen in AD~\cite{arvanitakis2019diagnosis}. Moreover, VaD often coexists with neurodegenerative pathology, resulting in mixed dementia and complicating both diagnosis and disease modeling~\cite{kalaria2016neuropathological}.  

Clinical diagnosis of VaD involves neurocognitive testing, vascular risk profiling, and neuroimaging. Established criteria such as those of the National Institute of Neurological Disorders and Stroke and the Association Internationale pour la Recherche et l’Enseignement en Neurosciences (NINDS-AIREN), along with DSM-5 guidelines, provide clinical frameworks; however, diagnosis increasingly relies on multimodal imaging. MRI is essential for detecting white matter hyperintensities, lacunar infarcts, strategic infarcts (e.g., in the thalamus or angular gyrus), and microbleeds, while advanced techniques such as DTI and arterial spin labeling (ASL) further support the identification of microvascular damage and chronic hypoperfusion~\cite{frantellizzi2020neuroimaging}. Neuropathologically, VaD manifests as a network disconnection syndrome due to cumulative microstructural damage across distributed brain systems. Key affected regions include the white matter, where leukoaraiosis and demyelination disrupt interregional communication; the basal ganglia, particularly in lacunar infarcts; the thalamus, which serves as a relay hub for cortical-subcortical communication; and the frontal lobes, which are crucial for attention and executive function. Hippocampal involvement is typically less prominent than in AD, though it may become more apparent in mixed pathology cases or when perfusion deficits extend into medial temporal regions~\cite{kalaria2016neuropathological}. This distinct spatial distribution of lesions, combined with vascular imaging biomarkers and cognitive profiles, supports differential diagnosis and the development of targeted intervention strategies for VaD.

\subsection{Frontotemporal Dementia}  

FTD is a group of neurodegenerative disorders primarily characterized by early deterioration in behavior, personality, or language. It is recognized as the leading cause of dementia in individuals under the age of 65. Unlike AD, which predominantly presents with memory impairment, FTD manifests with executive dysfunction, social disinhibition, apathy, or progressive aphasia, depending on the subtype involved~\cite{rohrer2013neuroimaging, convery2019clinical}. Major variants include the behavioral variant (bvFTD), nonfluent/agrammatic variant primary progressive aphasia (nfvPPA), and semantic variant PPA (svPPA)~\cite{bib89f}. The underlying pathology is typically associated with abnormal protein accumulations, such as tau or fused in sarcoma (FUS), leading to selective neuronal loss and gliosis~\cite{seelaar2011clinical}. FTD generally follows a progressive course, with marked declines in social cognition, language, and executive function. While certain subtypes may initially spare memory, widespread cognitive impairment often develops as the disease advances. In some cases, progression may involve motor neuron disease or extrapyramidal symptoms, further complicating clinical management~\cite{bang2015frontotemporal}.  

Diagnosis relies on detailed clinical history and behavioral assessment, supported by structural and functional neuroimaging. MRI typically reveals focal atrophy in the frontal and anterior temporal lobes, with patterns of asymmetry depending on the subtype. For example, bvFTD commonly affects the orbitofrontal cortex, insula, and anterior cingulate cortex—regions implicated in emotion regulation and social behavior. In contrast, svPPA is associated with pronounced left-sided anterior temporal lobe atrophy, impairing language comprehension, while nfvPPA involves degeneration of the left posterior frontal lobe and supplementary motor areas, disrupting speech production and grammatical processing~\cite{rohrer2013presymptomatic}. Some FTD variants also show subcortical involvement, particularly in the basal ganglia and thalamus, especially when motor symptoms or genetic mutations are present. The relative sparing of the posterior parietal and occipital cortices further distinguishes FTD from AD and other dementias. As disease-modifying treatments remain limited, early recognition and accurate subtyping based on neuroanatomical patterns are essential for prognosis, caregiver planning, and determining eligibility for clinical trials.

\subsection{Parkinson’s Disease Dementia}

PDD refers to the onset of dementia in individuals with a pre-existing diagnosis of PD, typically emerging after several years of motor symptoms such as bradykinesia, rigidity, and resting tremor. It is estimated that up to 80\% of individuals with PD will eventually develop dementia, making cognitive decline a major non-motor complication of the disease~\cite{emre2007clinical}. The underlying pathology of PDD is multifaceted, involving widespread neurodegeneration that extends beyond the classical dopaminergic system. While degeneration of dopaminergic neurons in the substantia nigra pars compacta accounts for motor symptoms, the spread of alpha-synuclein pathology (Lewy bodies) to limbic and neocortical regions is central to cognitive decline~\cite{fang2020cognition}. Unlike AD, which is primarily driven by beta-amyloid and tau pathology, PDD is considered a synucleinopathy, frequently accompanied by cholinergic and noradrenergic deficits. Clinically, PDD is characterised by a distinctive cognitive profile, with early impairments in executive function, attention, and visuospatial abilities. Memory deficits tend to occur later and are typically less pronounced in the initial stages. Neuropsychiatric symptoms such as hallucinations, apathy, and fluctuating cognition are also common and can complicate disease management~\cite{litvan2011diagnostic}.

Diagnosis is based on clinical criteria, such as those proposed by the Movement Disorder Society, which require the onset of dementia at least one year after PD diagnosis to differentiate PDD from DLB. Neuropsychological assessments focusing on frontal-subcortical domains, alongside MRI or PET imaging, aid in confirming the diagnosis and excluding other causes of cognitive decline. Neuroimaging and neuropathological studies have revealed consistent patterns of regional brain involvement in PDD. Early degeneration is observed in the substantia nigra and basal ganglia, particularly the putamen and caudate nucleus, disrupting dopaminergic circuits vital for motor and cognitive integration. The thalamus, which modulates cortico-subcortical communication, is also frequently affected. As the disease progresses, pathology spreads to the prefrontal cortex, impairing executive function and working memory. Degeneration of posterior cortical regions, including the parietal and occipital lobes, contributes to visuospatial dysfunction and hallucinations, which help distinguish PDD from other dementias such as FTD or VaD~\cite{burton2004cerebral}. The distributed involvement of both subcortical nuclei and heteromodal association cortices underscores PDD as a network-level disorder, necessitating multidimensional diagnostic and therapeutic strategies.

\begin{figure}[t]
    \centering
    \includegraphics[width=0.9\textwidth]{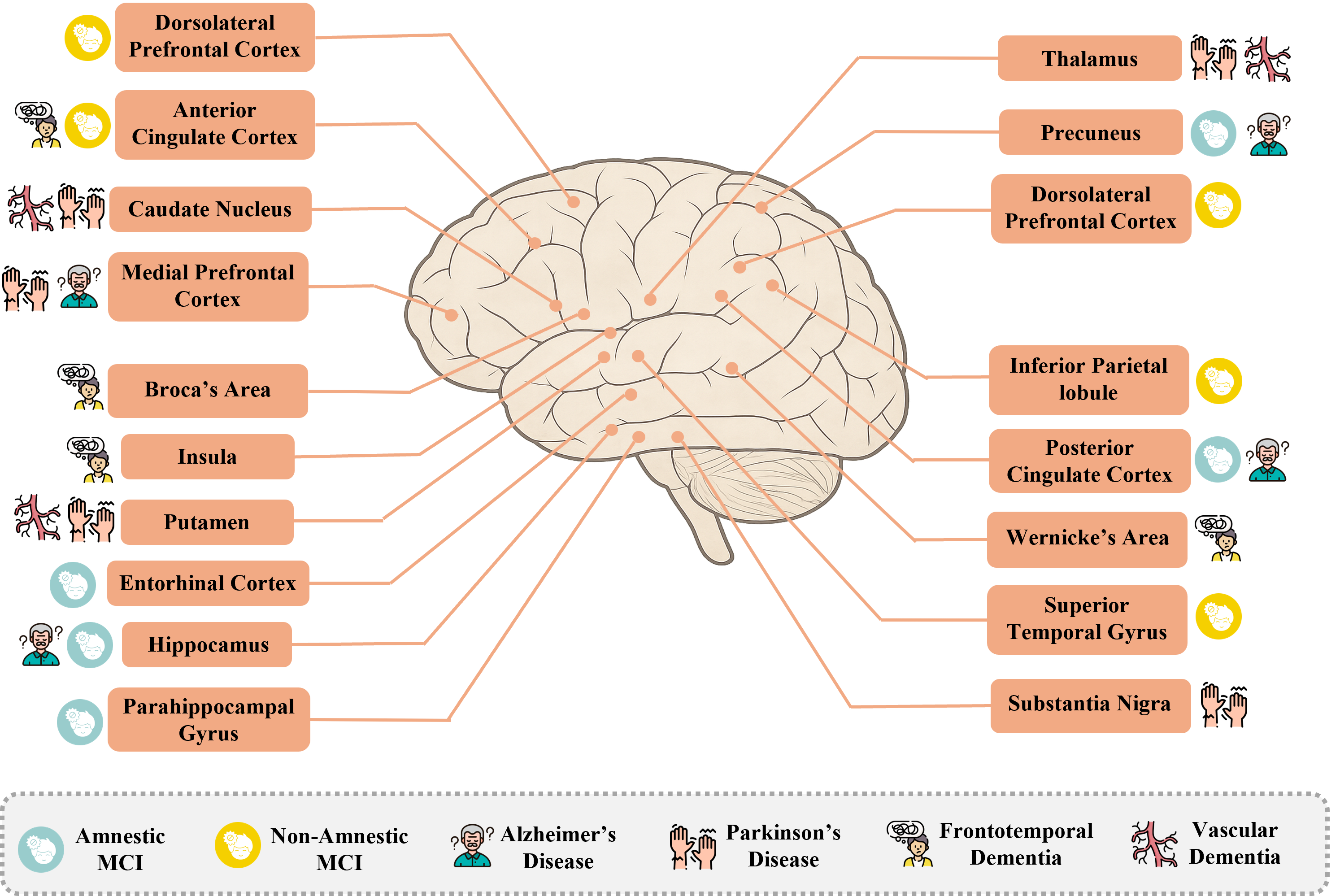} 
    \caption{Brain regions affected across dementia subtypes. Each icon represents a specific subtype of dementia, and icons adjacent to a brain region indicate that the corresponding condition impacts the region.}
    \Description{The affected brain regions}
    \label{fig: brain region}
\end{figure}

As dementia encompasses a spectrum of clinically and anatomically heterogeneous conditions, XAI has emerged as a critical tool for identifying disease-specific neural signatures and supporting transparent clinical decision-making. Although subtypes share overlapping symptoms such as cognitive decline, memory impairment, and behavioural disturbances, each is characterised by distinct patterns of brain region involvement. As illustrated in Figure~\ref{fig: brain region}, regions such as the posterior cingulate cortex, precuneus, and prefrontal cortex are implicated in multiple forms of dementia, including AD, VaD, and PDD, suggesting partially convergent neurodegenerative processes. In contrast, areas such as the insula and Broca’s area exhibit selective vulnerability in FTD, while subcortical structures like the substantia nigra and putamen are uniquely affected in PDD.

In this context, XAI provides a robust framework for bridging the gap between complex model decisions and clinical interpretability. By identifying the brain regions and connectivity patterns that underpin subtype-specific predictions, XAI facilitates deeper insights into disease mechanisms, enables more informed clinical decisions, and supports personalised diagnostic strategies. The coexistence of overlapping anatomical vulnerabilities and distinct regional signatures across dementia subtypes underscores the necessity for interpretable models capable of distinguishing between shared and unique biomarkers. Building on these challenges and opportunities, the following sections systematically examine recent advances in XGNN techniques applied to dementia research, with an emphasis on their role in elucidating both common and subtype-specific neuroanatomical patterns.

\begin{figure}[t]
    \centering
    \includegraphics[width=0.9\textwidth]{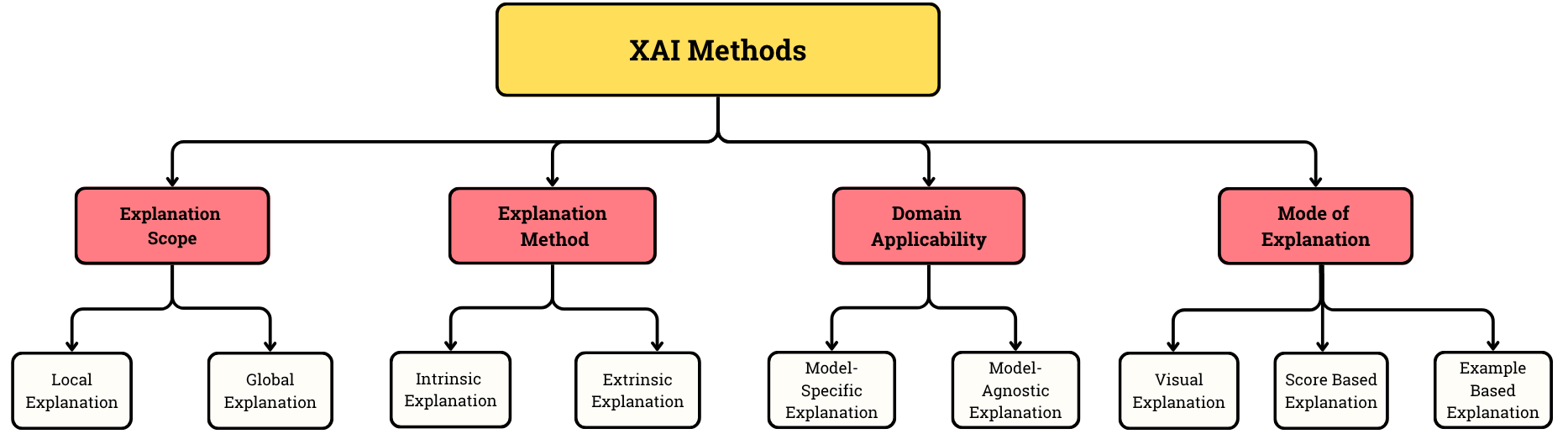} 
    \caption{The taxonomy of the XAI methods in dementia research.}
    \label{taxonomy}
    \Description{Taxonomy of XAI methods for dementia}
\end{figure}

\section{A Taxonomy of XGNN}
\label{sec:section4}

For brain graphs, the granularity of interpretation can occur at multiple levels: subject-specific, population, and feature level. At the subject-specific level, the brain is represented as a graph, where connections encode network disruptions associated with dementia. At the population level, each node represents an individual subject, while edges reflect inter-subject similarities or relationships, facilitating group-level analysis and subtype discovery. At the feature level, the graph represents features such as imaging biomarkers, cognitive scores, or genetic markers, with edges denoting correlations or dependencies, often derived from measures such as Pearson correlation or mutual information. This multilevel perspective enables the investigation of feature interactions and supports disease characterisation and network modelling.

Building on these representations, we introduce a taxonomy of XAI methods relevant to graph-based models in the domain of dementia research. As illustrated in Figure~\ref{taxonomy}, we categorise XAI approaches along four key dimensions: (1) \textit{Explanation Scope}, distinguishing between global and local explanations; (2) \textit{Explanation Method}, encompassing intrinsic and extrinsic techniques; (3) \textit{Domain Applicability}, differentiating between model-specific and model-agnostic approaches; and (4) \textit{Mode of Explanation}, referring to the modality of outputs, such as visualisations, example-based reasoning, or score-based metrics. While this framework builds upon previous work~\cite{van2022explainable, bib157, patricio2023explainable}, we extend existing taxonomies by explicitly incorporating score-based explanations as a distinct category. This addition is particularly relevant for identifying biomedical risk factors, especially in the context of dementia, where quantifiable interpretability can support both clinical insight and model accountability.

\subsection{XAI Methods}
\subsubsection{Self-Explanation}

While the terms \textit{interpretability} and \textit{explainability} are often used interchangeably~\cite{bib134sxai}, this study adopts the term \textit{Self-Explanation} (S-XAI) within the context of interpretability, as S-XAI methods are inherently designed to be interpretable~\cite{bib135sxai}. Unlike post-hoc techniques, S-XAI approaches elucidate the internal mechanisms of AI models and intrinsically provide insights into their decision-making processes, offering transparency on how input features influence predictions. Kakkad et al.~\cite{bib117xai} categorized self-explainable methods based on information and structural constraints, while Melis et al.~\cite{bib137sxai} refined the concept through the principles of fidelity, diversity, and grounding. Similarly, Velez et al.~\cite{bib121kx} introduced a framework for evaluating S-XAI methods using application-grounded, human-grounded, and functionality-grounded assessments. In the context of medical image analysis, Hou et al.~\cite{bib135sxai} further categorized S-XAI techniques into input, model, and output explainability. Overall, S-XAI plays a vital role in fostering trust and transparency in AI systems, particularly in high-stakes domains such as healthcare, by providing built-in and comprehensible explanations of model behavior.

\subsubsection{Post-hoc Explanation}

Post-hoc explanation methods are designed to interpret decisions of already trained models without modifying their architecture. These approaches analyse model predictions after training to reveal the underlying decision-making process, and they have been widely adopted in dementia research to highlight critical brain regions, features, and connectivity patterns.

\textit{Gradient-based methods} rely on gradients of the model output with respect to the input to identify influential features. Representative techniques include Gradient-weighted Class Activation Mapping (Grad-CAM)~\cite{selvaraju2020grad, zhou2016learning}, which highlights key regions or nodes contributing to predictions and has been adapted from CNNs to GNNs in cognitive disorder studies; saliency maps~\cite{bib141sa}, which generate visualisations of the most important input elements; and guided backpropagation (GBP)~\cite{springenberg2014striving, bib142gbp}, which refines saliency visualisations by suppressing irrelevant signals and emphasising positively contributing features.

\textit{Perturbation-based methods} are model-agnostic approaches that explain predictions by altering the input and observing changes in the output. Key techniques include SHAP (Shapley Additive Explanations)~\cite{lundberg2017unified}, which assigns game-theoretic importance values to features and has been applied to link imaging biomarkers and cognitive measures to dementia predictions, and GNNExplainer~\cite{ying2019gnnexplainer}, which is specifically designed for GNNs to identify explanatory subgraphs and node features driving model outputs.

Overall, both gradient-based~\cite{bib117xai, bib138pxai} and perturbation-based~\cite{bib117xai, bib144gbp} post-hoc methods play a crucial role in enhancing the interpretability of complex models. By clarifying which features or connectivity patterns influence predictions, they foster greater trust in AI-assisted dementia diagnosis and support more transparent clinical decision-making.

\subsubsection{Hybrid Explanations} 
Disease mechanisms in dementia are often complex and multifactorial, which increases the need for transparent and explainable graph models that are critical for clinical adoption. Hybrid explanation approaches are seen as promising solutions because they combine multiple XAI techniques, such as saliency maps, causal inference, and feature attribution, to capture complementary aspects of a model's behavior. These approaches help researchers gain clearer insights by linking affected regions to human-based explanations~\cite{bib34s} and identifying which regions contribute to model predictions. By integrating various XAI methods, hybrid explanations improve reliability and robustness, foster trust in AI systems, and provide deeper insights into underlying pathological processes.  

\subsubsection{Emerging Trends} 
While many studies have explained key terms and concepts of XAI~\cite{bib130emt, bib123kx}, more recent work has focused on classifying explanation quality based on the Co-12 properties: correctness, completeness, consistency, continuity, contrastivity, covariate complexity, compactness, composition, confidence, context, coherence, and controllability~\cite{bib131emt}. Schwalbe et al.~\cite{bib130emt} proposed a unified taxonomy of XAI methods by conducting a systematic survey and offering a cohesive classification of each concept. In addition, there has been an increasing emphasis on causality, which emphasizes cause-effect relationships rather than correlations. Causality provides a deeper understanding of how input data influence outcomes and yields insights into causal relations in dementia studies~\cite{bib132emt, bib133emt}. Although these are not fundamental XAI concepts, they play a crucial role in ensuring the effectiveness and adaptability of explainable models.

\subsection{Definition Formulation}
Constructing a brain graph begins with pre-processing and aligning MRI scans to a standard atlas for automated ROI segmentation, yielding a parcellation into $n$ discrete regions. These regions serve as the graph's nodes and define its spatial resolution. In this survey, we examine brain data at three complementary levels, subject-specific, population-level, and feature-level—each represented as an undirected graph $G^{(\ell)} = ( V^{(\ell)}, E^{(\ell)})$, with an associated adjacency matrix $A^{(\ell)} \in \mathbb{R}^{|V^{(\ell)}| \times |V^{(\ell)}|}$ and node feature matrix $X^{(\ell)} \in \mathbb{R}^{|V^{(\ell)}| \times d^{(\ell)}}$.

At the subject-specific level (\(\ell = s\)), \(V^{(s)} = \{v_i\}_{i=1}^n\) denotes anatomical or functional ROIs, and the adjacency matrix is defined as
\begin{equation}
    A_{ij}^{(s)} = 
    \begin{cases}
        w_{ij}, & \text{if } (v_i, v_j) \in E^{(s)} \\
        0, & \text{otherwise}
    \end{cases},
\end{equation}
where \(w_{ij}\) may represent tract strength from DTI or functional correlation from fMRI. Each node \(v_i\) is associated with a feature vector \(x_i \in \mathbb{R}^d\), such that \(X^{(s)}\) aggregates these row-wise.

At the population level (\(\ell = p\)), \(V^{(p)} = \{u_j\}_{j=1}^m\) represents individual subjects, and edges encode inter-subject similarity, for instance, via a Gaussian kernel on clinical measures. This yields an adjacency matrix \(A^{(p)} \in \mathbb{R}^{m \times m}\) and a subject feature matrix \(X^{(p)} \in \mathbb{R}^{m \times d'}\).

At the feature level (\(\ell = f\)), \(V^{(f)} = \{f_k\}_{k=1}^p\) corresponds to imaging biomarkers, cognitive scores, or genetic markers. Edges in \(E^{(f)}\) reflect statistical dependencies, such as Pearson correlation or mutual information, resulting in \(A^{(f)} \in \mathbb{R}^{p \times p}\), with feature-level attributes captured in \(X^{(f)} \in \mathbb{R}^{p \times d''}\).

To process these graphs, we employ GNNs, which define a parametric mapping
\begin{equation}
    f_{\theta} : (A, X) \mapsto Z,
\end{equation}
where \(Z\) denotes either node-level embeddings or graph-level predictions. In the message-passing paradigm, each GNN layer \(l = 0, \dots, L-1\) updates node embeddings as follows:
\begin{equation}
    h_v^{(l+1)} = \gamma^{(l)} \left(h_v^{(l)}, \bigoplus_{u \in \mathcal{N}(v)} \phi^{(l)}(h_v^{(l)}, h_u^{(l)}, A_{vu}) \right),
\end{equation}
where \(h_v^{(0)} = X_v\), \(\mathcal{N}(v) = \{ u : (u, v) \in E \}\) denotes the neighbourhood of \(v\), \(\phi^{(l)}\) is the learnable message function, \(\gamma^{(l)}\) is the update function, and \(\bigoplus\) is a permutation-invariant aggregator (e.g., sum, mean, or max). After \(L\) layers, the final node embeddings \(\{ h_v^{(L)} \}_{v \in V}\) are obtained. For graph-level outputs, a readout operation is applied:
\begin{equation}
    z = \text{READOUT}(\{ h_v^{(L)} \}).
\end{equation}

To enhance interpretability, we incorporate an explanation framework. Let \(\hat{y} = f_{\theta}(A, X)\) denote the model's output. An explainer is defined as a function
\begin{equation}
    \phi : (A, X, f_{\theta}, \hat{y}) \mapsto \mathcal{E},
\end{equation}
where the explanation object \(\mathcal{E}\) may include node importance scores \(S_V \in [0, 1]^{|V|}\), edge importance scores \(S_E \in [0, 1]^{|E|}\), or feature masks \(M \in [0, 1]^{|V| \times d}\).

\subsection{Explanation Scope}
The scope of explanation differentiates whether it addresses the reasoning behind specific predictions (\textit{local}) or provides insight into the behaviour of the overall model (\textit{global}).

\subsubsection{Global Explanation} 

Global explanation provides insights into a model’s decision-making process at the dataset level. Let $\mathcal{D} = \bigl\{(A_i, X_i, y_i)\bigr\}_{i=1}^N$ denote a dataset of \(N\) brain graphs at a given level \(\ell\), and let $f_{\theta} \colon (A, X) \mapsto \hat{y}$ be a trained GNN. A global explanation is defined as an object $E_{\mathrm{global}} \;\in\; \mathcal{E}_{\mathrm{global}}$, obtained via an explainer
\begin{equation}
    \phi_{\mathrm{global}} \colon (\mathcal{D}, f_{\theta}) \;\longrightarrow\; E_{\mathrm{global}}.
\end{equation}
In other words, \(E_{\mathrm{global}}\) is derived from both the entire dataset and the trained model, and provides a summarised understanding of model behaviour across all samples.

For instance, global explanation methods can identify brain regions that are consistently important across the dataset by computing average importance scores, rather than focusing on individual patients. One approach involves reporting the frequency of selected features or highlighting the top-ranked brain regions across the cohort, as demonstrated in~\cite{bib72bf}, which emphasises general patterns. Another example includes the use of average CAM to visualise brain regions and assess how each node, edge, or feature contributes to the predictions of \(f_{\theta}\) across all \(N\) samples~\cite{bib43s}.

\subsubsection{Local Explanation} 
Local explanation provides insights into individual patients and how a model arrives at its decision for a specific case. For a single graph instance \( G_i = (V_i, E_i, A_i, X_i) \) with model output \( \hat{y}_i = f_\theta(A_i, X_i) \), a local explanation is an object \( E_{\mathrm{local}}(i) \in \mathcal{E}_{\mathrm{local}}(i) \) produced by
\begin{equation}
    \phi_{\mathrm{local}} \colon (A_i, X_i, f_\theta, \hat{y}_i) \longrightarrow E_{\mathrm{local}}(i).
\end{equation}
Here, \( E_{\mathrm{local}}(i) \) may include per-node importance scores \( S_V(i) \), per-edge scores \( S_E(i) \), or a feature mask \( M(i) \), and serves to explain why \( f_\theta \) made the prediction \( \hat{y}_i \) on the individual graph \( G_i \). 

An example of a local explanation involves constructing subject-specific graphs to analyse discriminative information flows related to specific pathogenic factors~\cite{bib11}, or interpreting predictions by comparing a patient's features to those of others from different diagnostic categories~\cite{bib39s}.

\subsection{Explanation Method} 
The distinction between intrinsic and extrinsic explanations lies in whether interpretability mechanisms are integrated during model design (intrinsic) or applied post hoc after model training (extrinsic).

\subsubsection{Intrinsic Explanation} 
Intrinsic explanations, also known as \textit{ante-hoc} or model-based explanations, are designed to be inherently interpretable while producing human-understandable representations. For instance, attention weights derived from an attentive pooling mechanism are assigned to different relation types based on the model's internal structure~\cite{bib39s}. Another example involves the use of gradient information to generate heatmaps that highlight salient brain regions and specific time points, thereby providing intrinsic interpretability through the model’s own decision-making process~\cite{bib78bio}.

\subsubsection{Extrinsic Explanation} 
Extrinsic explanations refer to methods used to interpret and clarify the predictions of black-box models after training has been completed. For example, in one study~\cite{bib48dp}, GNNExplainer was applied post hoc to identify important brain ROIs, thereby serving as an extrinsic explanation. Another instance involves the use of gradient-based methods after model training to visualise salient brain regions~\cite{bib43s}.

\subsection{Domain Applicability} 

This category distinguishes between explanations that are tailored to a specific model architecture (specific) and those that are applicable across a range of models (agnostic).

\subsubsection{Model-Specific Explanation} These explanations are specific to individual models and their architectures, and therefore lack generalisability. For instance, in~\cite{bib153dp, bib65bf}, the explainability is inherently tied to the model architecture itself. In particular,~\cite{bib65bf} employs contrastive pooling with a dual attention mechanism to facilitate interpretability in brain classification tasks.

\subsubsection{Model-Agnostic Explanation} 
These explanations are universally applicable, regardless of the model’s internal architecture. They can be employed to interpret arbitrary models without being constrained to a specific model type. Shapley Additive Explanations are model-agnostic, allowing their application across a wide range of models~\cite{bib24s}.

\subsection{Mode of Explanation}
The mode of explanation refers to the type of explanation generated by an explanation method. Among the reviewed approaches, explanations are typically presented in three forms: visual (e.g., heatmaps, saliency maps, or CAM), cognitive score-based, and example-based. While these categories are conceptually distinct, they often overlap in practice. For instance, a visual explanation may be paired with an example-based explanation to illustrate how the model utilises specific brain regions in its prediction while providing a human-readable rationale~\cite{bib39s}. Similarly, score-based explanations can be integrated with visual approaches by highlighting high-scoring ROIs to support disease identification~\cite{bib153dp, bib55dp}.

\subsubsection{Visual explanation} 
An explanation method that employs visual techniques illustrates where, how, and what the model attends to when making decisions. For instance, the study in~\cite{bib153dp} projected clusters onto the cortical surface of the brain to highlight regions associated with AD. Similarly,~\cite{bib78bio} applied saliency heatmaps to reveal which brain areas were most relevant for distinguishing between MCI and healthy individuals.

\subsubsection{Score-based explanation} 
Scope-based explanations assign numerical values to features (i.e., brain regions) to reflect their importance or contribution to the model’s performance. In~\cite{bib153dp}, Grad-CAM was used to generate numerical importance values for each brain region, which were subsequently ranked to determine their potential as biomarkers. Similarly,~\cite{bib76bio} introduced a method in which importance probabilities were learned during model training to quantify each feature’s contribution to decision-making, thereby facilitating the identification of potential imaging and genetic biomarkers. In~\cite{bib24s}, SHAP scores were employed to evaluate feature informativeness for disease classification. Features such as education level and APOE status were assigned high importance scores, highlighting their roles as risk factors in the identification of AD.

\subsubsection{Example-based explanation} 
This type of explanation interprets the model's decision-making process through specific instances (examples), illustrating how the model responds to changes in input. For instance, \cite{bib39s} employs example-based explanations to elucidate the model's behaviour and demonstrate how it captures complex relationships between patients. Another application involves example-based explanations at the group level, highlighting significant regions or genes that contribute to disease classification~\cite{bib11}.

\section{Mild Cognitive Impairment}
\label{sec:section5}
\begin{figure}[t]
    \centering
    \includegraphics[width=1.0\textwidth]{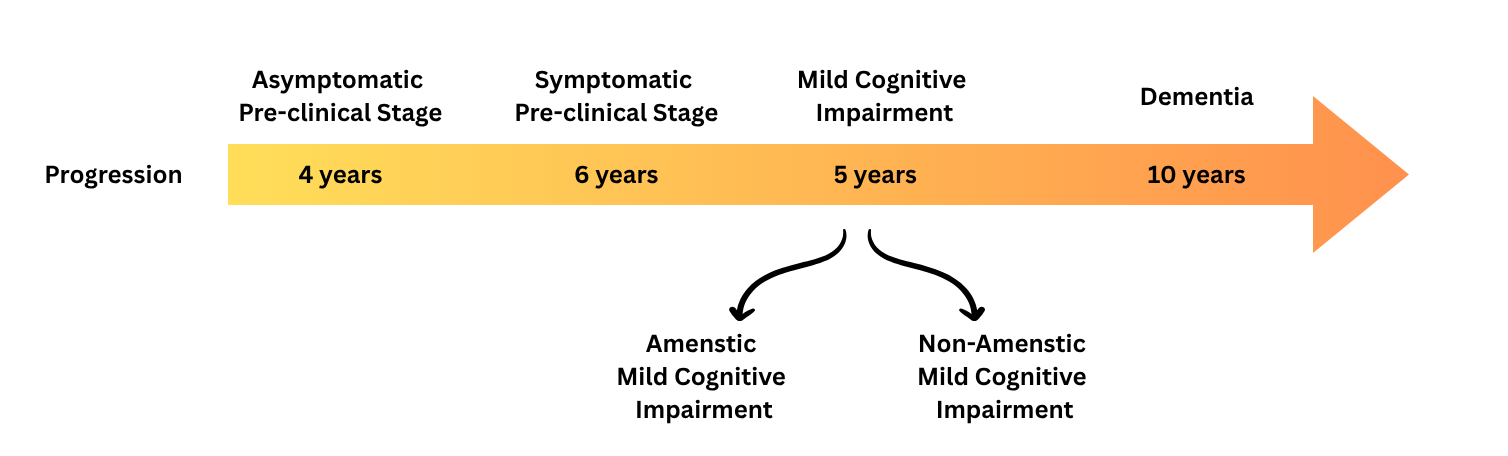} 
    \caption{Progression course from Mild Cognitive Impairment to Dementia.}
    \label{fig:preclinical}
    \Description{The figure illustrates the clinical progression from Mild 
    Cognitive Impairment to Dementia. It shows the preclinical stage, the 
    onset of mild symptoms, and the gradual decline leading to severe dementia.}
\end{figure}

Recent studies on MCI have increasingly focused on identifying predictive markers and developing early diagnostic frameworks to account for its heterogeneous progression. While clinically recognized as a prodromal stage, MCI encompasses a wide spectrum of cognitive presentations and trajectories. This variability has led to a growing body of longitudinal studies exploring individual-level factors, such as baseline cognitive profiles, neuroimaging biomarkers, and genetic predisposition, that influence conversion rates to dementia~\cite{bib159}. Figure~\ref{fig:preclinical} illustrates the clinical progression from MCI to dementia, highlighting the preclinical stage, the onset of mild symptoms, and the gradual decline leading to severe dementia.

Advanced neuroimaging techniques, including sMRI, fMRI, and PET imaging, have been employed to investigate network-level dysfunctions and regional brain changes associated with both aMCI and naMCI subtypes. These studies suggest that alterations in connectivity patterns, particularly within the default mode, salience, and executive control networks, may serve as early indicators of disease progression~\cite{bib160, bib161}. The integration of imaging data with clinical and neuropsychological assessments is therefore critical for improving diagnostic accuracy and patient stratification.

Moreover, data-driven methods such as graph-based models have emerged as promising tools for capturing the subtle and complex brain changes associated with MCI. Such approaches enable the modeling of inter-regional connectivity and can aid in distinguishing between stable MCI, progressive MCI, and normal aging. Importantly, explainable models are beginning to shed light on which features or brain regions are most relevant for predicting outcomes, thereby contributing to personalized and targeted intervention strategies. A summary of key studies incorporating multimodal data and computational frameworks for MCI classification is presented in Table~\ref{tab:MCI_table}.

\begin{table}[t]
\centering
    \caption{Research work related to MCI diagnosis and prediction.}
    \scriptsize
    \begin{tabular}{p{1.5cm}p{2.0cm}p{2.0cm}p{1.0cm}p{2.5cm}p{2.0cm}p{2.3cm}}
    \toprule
    \textbf{XAI Method} & \textbf{Architecture} & 
    \textbf{Modality} & \textbf{Level} & 
    \textbf{Nature of Explanation} & \textbf{Performance} & 
    \textbf{Application} \\
    \midrule
    \multicolumn{7}{c}{\textbf{Amnestic MCI}} \\
    \midrule
    \multirow{8}{*}{\parbox{1.5cm}{\raggedright Post-Hoc Explainable}}
    &GCN-GradCAM & MRI, fMRI & Sub & Both/In/Sp/Vis & 
    ACC: 89.16\%, 92.45\% & Early diag., AD pathology~\cite{bib78bio} \\
    &GATConv- \newline GradCAM & MRI & Sub & Both/In/Sp/Vis & 
    ACC: 82.6\%  & Brain shape staging, \newline highest seen on eMCI vs lMCI~\cite{bib29s} \\
    &GCN-GradCAM & fMRI & Sub & Both/In/Sp/Vis & 
    ACC: 92.2\% & MCI Diagnostic accuracy~\cite{bib57bf} \\
    &GCN-GradCAM & fMRI & Both & Both/Ex/Sp/Vis+Exmpl & 
    ACC: 78.6\% & FC-EC analysis~\cite{chen2025guiding} \\
    & GCN-Saliency & fMRI, SNP & Popln & Both/In/Sp/Vis+Score & 
    ACC: 89.4\%, 85\%, \newline 86.4\% & AD/MCI \newline subtype clsfn~\cite{bib81bio} \\
    & GCN-\newline GNNExplainer & MRI, PET & Popln & Both/In/Ag/Vis & 
    AUC: 88.51\% & Focus on \newline A$\beta$ prediction~\cite{bib79bio} \\
    & GNN-\newline GNNExplainer & fMRI & Sub & Both/Ex/Ag/Vis & 
    MAE: 5.92 ± 0.62 & Brain age gap (BAG), subtype clsfn~\cite{bib48dp} \\
    & GNN- \newline GNNExplainer & MRI & Popln & Both/In/Sp/Score & 
    ACC: 94.4\%, 87.5\% & MCI subtype and conversion~\cite{bib42s} \\
    \midrule
    \multirow{6}{*}{\parbox{1.5cm}{\raggedright Self Explainable}} 
    & GCN-Built-In & MRI & Popln & Both/In/Sp/Vis+Score & 
    ACC: 79.4\%, 68.4\% & AD-MCI diag. and MCI subtype clsfn~\cite{bib41s} \\
    & GNN-Built-In & MRI, fMRI, PET & Sub & Both/In/Sp/Vis+Exmpl & 
    $R^2$: 0.214 ± 0.08 & Early AD detectn.~\cite{youn2024brain} \\
    & GNN-Attention & MRI, DWI, PET & Popln & Gl/In/Sp/Vis & 
    ACC: 68\% & Group discriminative \newline pattern observed~\cite{bib66bf} \\
    & GCN-Attention & fMRI & Sub & Both/In/Sp/Vis & 
    ACC: 71.2\% & MCI biomarker \newline discovery~\cite{bib152dd} \\
    & GCN-Attention & fMRI, SNP & Popln & Gl/In/Sp/Vis+Score & 
    ACC: 91.5\%, 88.3\% \newline (MCI vs HC), (brain-gene) & 
    Multi-view AD \newline pathogen mapping~\cite{bib32s} \\
    & GCN-Attention & MRI, fMRI, genetic & Sub & Both/In/Sp/All & 
    ACC: 84.4\% & MCIn and MCIp prediction~\cite{bib11} \\
    \midrule
    \multicolumn{7}{c}{\textbf{Non-Amnestic MCI}} \\
    \midrule
    {\parbox{1.2cm}{\raggedright Post-Hoc Explainable}}
    & GCN-GradCAM & fMRI & Sub & Both/In/Sp/Vis & 
    ACC: 80.4\%, 79\% \newline (nMCI/aMCI) & Early VaD diag.~\cite{bib38s} \\
    \bottomrule
    \end{tabular}
    \caption*{\scriptsize \textbf{Level}: Sub = Subject, Popln = Population. \textbf{Nature of Explanation}: Scope / Method / Domain / Mode. Scope: Lo = Local, Gl = Global, Both; Method: In = Intrinsic, Ex = Extrinsic; Domain: Sp = Model Specific, Ag = Agnostic; Mode: Vis = Visual, Score = Numeric, Exmpl = Example. clsfn.=classification}
    \label{tab:MCI_table}
\end{table}

\subsection{Non-Amnestic MCI}
In the context of naMCI, Liu et al.~\cite{bib38s} proposed a multi-scale atlas-based GCN to predict cognitive outcomes in individuals with subcortical ischaemic vascular disease (SIVD), a recognized precursor of vascular dementia (VaD). The model constructs individualized brain networks from fMRI data and integrates multi-scale brain parcellation to capture hierarchical representations of brain connectivity, achieving an accuracy of 80.4\% in differentiating naMCI from aMCI. To enhance interpretability, the authors employed GradCAM, a widely used post-hoc explainability technique, which provided intuitive visual explanations. Their analysis revealed that the limbic network was a key predictor for aMCI in the context of SIVD, whereas the salience and default mode networks were associated with non-cognitive impairment. These findings align with existing evidence linking vascular cognitive impairment to disruptions in the limbic network, underscoring both the model’s explainability and its clinical relevance in supporting VaD diagnosis.

\subsection{Amnestic MCI}

\subsubsection{Post-Hoc Explanation}
Studies have extensively focused on aMCI using XGNN techniques, employing both gradient-based and perturbation-based post-hoc methods. Zhang et al.~\cite{bib78bio} applied a spatiotemporal GCN combined with GradCAM on fMRI data to capture dynamic interactions across brain regions, highlighting critical areas such as the paracentral lobule, inferior occipital gyrus, and superior temporal gyrus. Guo et al.~\cite{bib29s} introduced a region-dependent graph attention convolution on brain morphable meshes, achieving an accuracy of 82.6\% in distinguishing early from late MCI. GradCAM visualizations emphasized subcortical structures such as the hippocampus, amygdala, and thalamus as pivotal for classification. He et al.~\cite{bib57bf} proposed a multimodal GCN integrating resting-state fMRI data with predictive coding, Granger causality, and sparse representation, reaching 92.2\% accuracy and an AUC of 0.98 on the ADNI dataset, and identifying the precuneus, supplementary motor cortex, and cuneus as key regions.

Saliency-based approaches have also been employed to classify MCI and elucidate the pathological basis of AD and its subtypes. Bi et al.~\cite{bib81bio} developed a feature aggregation GCN to model brain–gene interactions via node-to-node feature aggregation. Saliency analysis highlighted the hippocampus, superior and middle occipital gyri as critical regions, and identified LRP1B and CNTN5 as key genes associated with AD diagnosis. Built-in interpretability has further gained attention; for example, Zeng et al.~\cite{bib41s} designed a two-phase GCN that addressed training set bias and provided real-time interpretability, although with modest accuracy (68.4\%) in distinguishing stable (MCIn) from progressive MCI (MCIp).

GNNExplainer has been widely adopted to enhance interpretability in early AD and MCI studies. Kim et al.~\cite{bib79bio} applied GNNExplainer to a GCN integrating demographic, genetic, and neuroimaging features, predicting A$\beta$ positivity with an AUC of 0.86. A$\beta$ positivity, a crucial biomarker in AD, was linked to specific subgraphs and node attributes, with cortical thickness and APOE-$\varepsilon$4 status identified as dominant features in cognitively unimpaired individuals. Importantly, biomarker contributions varied across age groups: older cohorts prioritized amyloid PET uptake, whereas younger groups relied more on FDG-PET metabolic rates. Song et al.~\cite{bib42s} proposed an auto-metric GNN using genetic and multimodal imaging data, achieving 94\% accuracy in AD diagnosis and 87\% in predicting MCI conversion. GNNExplainer highlighted the hippocampus, entorhinal cortex, and middle temporal lobe as key discriminative regions. Similarly, Gao et al.~\cite{bib48dp} applied GNNExplainer on a GNN trained with ADNI data, identifying alterations in the hippocampus, amygdala, and parahippocampal gyrus.

\subsubsection{Self Explanation} 
Self-explainable methods are designed such that interpretability is inherently embedded within the model architecture. These models generate both predictions and explanations during inference, typically by enforcing constraints such as attention mechanisms, disentangled representations, or sparse message passing. In the context of MCI and early AD, attention-based mechanisms have been particularly effective in enhancing classification performance while providing insights into disease-relevant brain regions.

Cai et al.~\cite{bib66bf} proposed a hypergraph GNN that integrated amyloid-PET and FDG-PET imaging to capture higher-order interactions. Qualitative analysis of hyperedges revealed recurrent propagation patterns involving key regions such as the hippocampus and parahippocampal gyrus, consistent with established AD pathology. Ma et al.~\cite{bib152dd} introduced a multi-graph cross-attention-based, region-aware fusion network that dynamically weighted heterogeneous features from multiple modalities, improving interpretability and highlighting critical temporal, frontal, and parietal regions. Bi et al.~\cite{bib32s} advanced AD diagnosis through imaging-genetics fusion, achieving 91.5\% accuracy for MCI vs. HC classification and 88.3\% for brain–gene associations, with the hippocampus and entorhinal cortex identified as key biomarkers. Shang et al.~\cite{bib11} further developed a graph capsule convolutional network (GCCN) to predict MCI-to-AD conversion, where capsule mechanisms captured hierarchical relationships between imaging and genetic features, supporting more effective early intervention strategies.

\subsection{Discussion}
Early diagnosis remains a central priority in dementia research, as timely detection enables more effective clinical management and intervention strategies. Gradient-based methods such as GradCAM and model-agnostic techniques like GNNExplainer have been widely applied to aMCI, offering intuitive visual explanations and highlighting disease-relevant brain regions. In parallel, S-XAI methods demonstrate strong potential in leveraging longitudinal and multimodal data to provide inherently interpretable insights. Nevertheless, naMCI remains comparatively underexplored, highlighting the need for more targeted models that address the heterogeneity of clinically relevant subtypes. Future research directions should include advancing multimodal fusion frameworks, improving generalisability across diverse cohorts, and establishing standardised benchmarks for fair evaluation. Moreover, the refinement of explainability techniques, with validation from clinical domain experts, is essential to ensure not only methodological robustness but also trustworthiness and applicability in real-world healthcare settings.

\section{Alzheimer's Disease}
\label{sec:section6}
AD is the most common cause of dementia, accounting for more than half of all cases and predominantly affecting individuals over the age of 60. It is clinically characterised by progressive cognitive decline, memory impairment, and behavioural disturbances~\cite{bib85a,xie2024multimodal}. As the disease advances, patients gradually lose independence and become increasingly reliant on assistance with daily activities. Although no curative treatment exists, pharmacological interventions can alleviate symptoms and provide temporary relief~\cite{bib86a}. In the context of AD diagnosis and progression monitoring, recent advances have employed XGNN models to improve both diagnostic accuracy and interpretability. These models enable the identification of disease-relevant brain regions and biomarkers, offering greater transparency in clinical decision-making. Table~\ref{tab:AD_table} summarises representative studies that integrate neuroimaging, clinical, demographic, and genetic data into graph-based frameworks, alongside XAI techniques, for AD classification and prediction. For clarity, explainable GNN approaches for AD are organised into three main categories: \textit{post-hoc} methods, self-explainable methods, and hybrid approaches. This taxonomy provides a structured basis for evaluating how different explanation strategies contribute to enhancing the reliability and clinical utility of AD diagnosis.

\begin{table}
\centering
    \caption{Research work related to AD diagnosis and prediction.}
    \scriptsize
    \begin{tabular}{p{1.5cm}p{2.0cm}p{2.0cm}p{1.0cm}p{2.5cm}p{2.0cm}p{2.3cm}}
    \toprule
    \textbf{XAI Method} & \textbf{Architecture} & \textbf{Modality} & \textbf{Level} & \textbf{Nature of Explanation} & \textbf{Performance} & \textbf{Application} \\
    \midrule
    \multirow{8}{*}{\parbox{1.5cm}{\raggedright Post-Hoc Explainable}} 
    & GCN-GradCAM & MRI & Sub & Gl/Ex/Sp/Vis & 
    ACC: 96.35\% & ROI visualization on \newline meshes~\cite{bib43s} \\
    & GCN-GradCAM & VBM-MRI, PETs & Both & Both/Both/Sp/Vis+Score & 
    ACC: 81.8\%, \newline PC: 0.941 & Biomarkers, MMSE,\newline ADAS13~\cite{bib37s} \\
    & GCN-GradCAM & fMRI & Both & Both/Both/Sp/Vis+Score & 
    ACC: 88.6\%; 79\% & FC biomarkers~\cite{bib68bf} \\
    & GCN-GradCAM & MRI, fMRI & Both & Both/Both/Sp/Vis+Score & 
    ACC: 96\%  & Progression \newline prediction~\cite{bib21s} \\
    & GCN-GBP & sMRI, dMRI,\newline fMRI & Sub & Both/Both/Sp/Vis+Score & 
    ACC: 76.2 ± 4.0\% & A$\beta$ status \newline (A$\beta$-, A$\beta$+)~\cite{bib22s} \\
    & GCN- \newline GNNExplainer & DWI, fMRI & Sub & 
    Both/Both/Sp/Vis+Score & 
    ACC: 88\% & AD vs SCI, \newline Disease staging~\cite{bib12} \\
    & GIN-SHAP & fMRI & Both & Both/Both/Ag/Score & 
    ACC: 90.44\% & FC \& gender clsfn~\cite{bib30s} \\
    & GCN-SHAP & MRI, PET, demo. & Both & Both/Both/Sp/Vis+Score & 
    ACC: 96.70\% & Feature selection~\cite{bib24s} \\
    \midrule
    \multirow{20}{*}{\parbox{1.5cm}{\raggedright Self Explainable}} 
    & GNN (VNN)- \newline Built-In & MRI & Both & Both/In/Sp/Vis+Score & 
    MAE: 5.44 ± 0.18; \newline PC: 0.47 ± 0.07 & Brain age \newline prediction~\cite{bib50dp} \\
    & GNN-Built-In & fMRI & Both & Both/In/Sp/Vis+Score & 
    Precision: 0.79 & Brain disorder \newline diagnosis~\cite{bib67bf} \\
    & GCN-Built-In & EEG & Popln & Both/In/Sp/Vis & 
    EC: 89.1\%, \newline EO: 85.4\%, \newline EO+EC: 81.8\% & 
    EEG-based \newline AD detection and \newline graph clsfn.~\cite{bib36s} \\
    & GNN-LSTM \newline -Attention & MRI & Sub & Both/In/Sp/Vis+Score & 
    ACC: 93.67\% & Longitudinal AD \newline diagnosis~\cite{bib55dp} \\
    & GT-Attention & MRI, DTI & Sub & Both/In/Sp/Vis+Score & 
    MAE: 2.71 ± 0.07;\newline ACC: 85.9\% & Estimation of \newline Brain age ~\cite{bib52dp} \\
    & GCN-Attention & MRI & Feature & Both/In/Sp/Vis+Score & 
    ACC: 86.9\% & AD diagnosis~\cite{bib70bf} \\
    & GCN-Attention & MRI, \newline FDG-PET, AV45 & Sub & Both/In/Sp/Vis+Score & 
    ACC: 92.3 ± 2.1\% & AD clsfn.~\cite{bib40s} \\
    & GNN-Attention & MRI, PET, gene, \newline clinical, bio, demo. & Both & Both/In/Sp/Vis+Score & ACC: 94.2\% & AD clsfn.~\cite{bib64bf} \\
    & GCN-Attention & MRI & Both & Both/In/Sp/Vis & 
    ACC: 93.90\% & AD diagnosis~\cite{bib31s} \\
    & GCN-Attention & MRI & Sub & Both/Both/Sp/Vis & 
    ACC: 90.73\% & AD diagnosis~\cite{bib75bio} \\
    & GT-Attention & fMRI & Sub & Both/Both/Sp/Scores & 
    ACC: 92.31\% & AD diagnosis~\cite{bib25s} \\
    & GT-Attention & fMRI & Both & Both/Both/Sp/Vis+Score & 
    ACC: 88.2\%  & AD diagnosis~\cite{bib59bf} \\
    & GCN-Attention & MRI, PET, SNP & Both & Both/Both/Sp/Vis+Score & 
    PC: 0.829 (ADAS13), 0.771 (MMSE), 0.768 (Tau) & 
    AD diagnosis~\cite{bib76bio} \\
    & GCN-Attention & MRI & Both & Both/Both/Sp/Vis+Score & 
    ACC: 86.2\% , 74.2\% & AD diagnosis~\cite{bib26s} \\
    & GNN-Attention & MRI & Popln & Both/In/Sp/Vis+Exmpl & 
    Macro-F1: \newline 0.205–0.851, Micro-F1: 0.686–0.858 & 
    Multimodal \newline detection~\cite{bib39s} \\
    & GCN-Instance level & MRI & Sub & Both/In/Sp/Vis & 
    ACC: 84.4±0.36\% & AD diagnosis~\cite{bib71bf} \\
    & GCN-Feature level & MRI & Sub & Gl/In/Sp/Vis+Score & 
    ACC $\geq$ 85\% & AD diagnosis~\cite{bib72bf} \\
    & GCN-Feature level & MRI & Both & Both/In/Sp/Vis+Score & 
    ACC: 95.6\% (ADNI), \newline 97.5\% (AIBL), \newline 93.3\% (OASIS), 100\% (MIRIAD) & AD diagnosis \newline and prognosis~\cite{bib82bio} \\
    & GCN-Feature level & fMRI, SNP & Both & Both/Both/Sp/Vis+Score & 
    ACC: 93.61\% & AD diagnosis, \newline pathogen \newline identification~\cite{bib33s} \\
    & GraphSAGE- \newline Feature level & MRI & Sub & Both/Both/Sp/Vis & 
    ACC: 92.1\%, 91.4\% \newline (Intra/Inter CV) & AD Diagnosis~\cite{bib28s} \\
    \midrule
    \multirow{3}{*}{\parbox{1.5cm}{\raggedright Hybrid Explanation}}
    & GCN-\newline Counterfactual \& \newline Causal & PET & 
    Both & Both/In/Sp/Vis+Score & ACC: 84\%  & Causal inference~\cite{bib153dp} \\
    & GNN-Saliency \newline \& CAM & MRI, SNP, CSF & Both & Both/Both/Sp/Vis+Score & ACC: 87.6\% & AD Diagnosis~\cite{bib77bio} \\
    & GCN-\newline Decomposition \& \newline SHAP &
    MRI, PET, \newline demo., cognitive,\newline genetic & 
    Both & Both/Both/Both/\newline Vis+Score & 
    Correctness: 71\% & Human validated \newline explanations~\cite{bib34s} \\
    \bottomrule
    \end{tabular} 
\label{tab:AD_table}
\end{table}

\subsection{Post-Hoc Explanation}
Post-hoc explainable methods aim to interpret the predictions of trained GNN models after learning is complete. These approaches treat the model as a black box and apply external techniques, such as feature attribution (e.g., GNNExplainer), node importance ranking, or visualisation—to identify the input features, nodes, or subgraphs that most influenced a specific prediction. In the context of AD, post-hoc methods help clinicians understand which brain regions or connectivity patterns are critical for diagnosis without altering the underlying GNN architecture.

Several gradient-based methods have been employed to interpret brain connectivity in AD. Grad-CAM has been widely used with GCNs to highlight salient brain regions, although implementations vary across imaging modalities, graph structures, and architectures. For example, Azcona et al.~\cite{bib43s} applied Grad-CAM to sMRI-based GCNs using cortical and subcortical triangular meshes, identifying atrophy in the inferior parietal lobule and the intermediate sulcus of Jensen. Zhang et al.~\cite{bib21s} constructed FC networks from rs-fMRI and highlighted the DMN, sensorimotor network (SN), and visual network (VN) as biomarkers, alongside topological features mediating amyloid-$\beta$ and glucose metabolism. Liu et al.~\cite{bib68bf} proposed a hierarchical GCN based on multiscale FC networks from fMRI, identifying the LIM and DMN as discriminative regions. Zhou et al.~\cite{bib37s} integrated multimodal imaging into an interpretable GCN framework for AD classification, with Grad-CAM emphasising the putamen and pallidum as biomarkers associated with cognitive scores such as the MMSE and Pearson correlation.

Guided backpropagation (GBP) has also been applied to highlight positively contributing input features by modifying gradient flow. Dolci et al.~\cite{bib22s} combined sMRI, fMRI, and dMRI within a multimodal GCN to classify amyloid-$\beta$ (A$\beta$) status, achieving 76.2\% accuracy. GBP identified the hippocampus and thalamus as relevant to amyloid deposition, while highlighting modality-specific networks, including the default mode, cingulo-opercular, visual, and somatomotor networks (fMRI), as well as the frontal, parietal, and temporal lobes (dMRI). These results underscore the value of modality-aware interpretation for uncovering unknown A$\beta$ deposition patterns.

Perturbation-based methods such as SHAP have provided both global and local interpretability in AD studies, explaining node-level and clinical predictors. Wang et al.~\cite{bib30s} applied SHAP within a dual multi-task GIN (DMT-GIN), achieving 90\% accuracy and identifying the precentral and middle frontal gyri as key regions. Zhang et al.~\cite{bib24s} used SHAP with an auto-fusion GCN, reaching 97.8\% accuracy (HC vs. AD) and 96.56\% (AD vs. MCI vs. HC), and identifying education and the Cogstate Brief Battery as the most influential predictors.

GNNExplainer has been employed to enhance interpretability by extracting subgraphs and node features that drive predictions. Gamgam et al.~\cite{bib12} applied GNNExplainer to a Siamese GCN using dMRI and rs-fMRI, identifying cortical regions contributing to structural and functional disruptions in AD progression, particularly in subjective cognitive impairment (SCI). Although GNNExplainer is model-agnostic, its use within a Siamese GCN highlights how tailored architectures can improve the clinical relevance of extracted explanations.

In summary, post-hoc methods are flexible and widely applicable, as they can be used with any pre-trained GNN without modifying its architecture. They are especially valuable when interpretability is needed after deployment. However, because they infer explanations indirectly, their outputs may not faithfully reflect a model’s internal reasoning~\cite{madsen2022post}, raising concerns about misleading or inconsistent insights in high-stakes domains such as AD diagnosis.

\subsection{Self Explanation}
In the context of AD diagnosis, self-explainable GNNs highlight key brain sub-networks through interpretable attention weights, providing direct insight into disease biomarkers as part of the predictive process. This built-in interpretability is crucial for fostering trust among clinical experts, as it intrinsically validates and explains model outputs.

Several studies have introduced self-explainable architectures tailored to AD. Sihag et al.~\cite{bib50dp} proposed an explanation-driven covariance neural network (VNN), operating on sample covariance matrices of cortical thickness features to predict brain age. Tong et al.~\cite{bib67bf} integrated SC-FC brain network features with transfer learning to analyse FC impairments, while Klepl et al.~\cite{bib36s} employed a gated GCN using EEG power spectral density, with graph structure learning and pooling modules for interpretable AD prediction. Attention mechanisms, first introduced by Vaswani et al.~\cite{vaswani2017attention}, have been widely adopted to learn the importance of brain connections. For example, Cai et al.~\cite{bib52dp} applied a graph transformer using multimodal MRI for brain age prediction, incorporating attention scores to assess modality influence. Xiao et al.~\cite{bib70bf} developed a dual graph convolutional network that achieved over 80\% diagnostic accuracy while identifying the hippocampus and frontal white matter as biomarkers. Similarly, Kim et al.~\cite{bib39s} introduced HetMed, a heterogeneous graph learning model with attentive pooling that highlighted clinically relevant variables such as cognitive scores, reaching Macro-F1 scores of 0.774 and 0.813 on ADNI and OASIS-3 datasets.

Other interpretable approaches extend beyond attention-based mechanisms. Sparse GCNs~\cite{bib40s}, attention-based fusion models~\cite{bib64bf}, and spatiotemporal architectures~\cite{bib25s, bib75bio} have demonstrated improved feature extraction and highlighted key regions such as the hippocampus, precuneus, and amygdala. Song and Yoshida~\cite{bib59bf} developed a temporal graphormer that captured disease-related FC patterns with enhanced interpretability. At the same time, studies by Bi et al.~\cite{bib33s}, Zhu et al.~\cite{bib72bf}, Nguyen et al.~\cite{bib82bio}, and Song et al.~\cite{bib28s} advanced feature- and instance-level explanations, uncovering meaningful imaging–genetic associations and providing patient-specific feature attribution. Li et al.~\cite{bib71bf} further extended this by ranking influential features across cohorts to support personalised diagnostic insights.

In summary, self-explainable methods embed interpretability directly into the model architecture, offering more transparent and faithful explanations compared to post-hoc techniques. This integration not only fosters clinical trust but may also improve generalisability by incorporating sparsity or attention constraints~\cite{gao2024going}. However, designing such models remains challenging, often involving trade-offs in model capacity and flexibility~\cite{longa2025explaining}.

\subsection{Hybrid Explanation}
Hybrid explanations have been increasingly employed to enhance trust and robustness in Alzheimer’s disease (AD) research by integrating multiple XAI techniques and addressing distinct aspects of disease progression. These methods combine built-in interpretability features with external post-hoc explanations, thereby improving transparency and validating model behaviour. In practice, hybrid GNNs often leverage attention mechanisms during training while applying additional explanation tools such as saliency or SHAP after training, providing both intrinsic and extrinsic interpretability.

Several studies have demonstrated the value of hybrid approaches. Lei et al.~\cite{bib77bio} combined saliency and CAM within a dual multilevel GNN using sMRI, genetic, and protein data, achieving 87.6\% accuracy. Their model identified the inferior temporal gyrus, hippocampus, and amygdala as key regions, while canonical correlation analysis revealed risk factor correlations between 0.801 and 0.821. Tekkesinoglu and Pudas et al.~\cite{bib34s} applied decomposition-based XAI to a GCN framework, integrating neurocognitive, genetic, and brain atrophy markers. Their approach outperformed SHAP in individual patient classification, with 71\% expert validation and interpretability ratings above six. Dai et al.~\cite{bib153dp} proposed Graph-VCNet, which integrates GCNs with counterfactual causal inference on sMRI data to investigate the impact of A$\beta$ accumulation. Their model achieved 84\% classification accuracy while enabling personalised treatment prediction and providing causal insights into disease pathways.

Overall, hybrid methods balance interpretability and performance by combining intrinsic explainability with post-hoc interpretive tools. They provide more comprehensive insights into model decisions and are particularly valuable for complex medical applications. However, they also inherit limitations from both paradigms, including increased model complexity and computational overhead~\cite{li2025can}. Reconciling potentially conflicting explanation outputs further introduces challenges related to consistency and interpretive reliability.

\subsection{Discussion}
Research on AD has demonstrated the effectiveness of XGNNs in identifying biomarkers and dysfunctional brain connectivity. Among post-hoc methods, SHAP and Grad-CAM are the most widely adopted, offering flexibility and the ability to explain GNN predictions in both spatial and feature domains. Attention-based approaches also provide effective visual explanations with competitive diagnostic performance. Hybrid methods, by combining intrinsic and extrinsic techniques, show strong potential for developing clinically grounded and human-understandable pipelines; however, only one study~\cite{bib34s} has validated explanations using correctness as a formal evaluation metric.

A key limitation in the current literature is the lack of consistent evaluation measures. Most studies report conventional performance metrics such as accuracy or AUC, but these do not capture the clinical interpretability or quality of the generated explanations. To advance the field, the selection of XAI techniques should be closely aligned with task-specific objectives, whether the focus is on visualizing ROIs, identifying biomarkers, modeling population-level patterns, or supporting personalized diagnosis. Integrating multiple explanation methods and adopting standardized evaluation metrics for explanation quality will yield more trustworthy insights and facilitate broader clinical adoption.

\section{Parkinson's Disease}
\label{sec:section7}
Building on the clinical understanding of PDD, recent studies have increasingly investigated the disorder through neuroimaging and computational modelling. Given the distributed network-level degeneration observed in PDD, graph-based representations of brain connectivity provide a compelling framework for disease characterization and diagnosis. These methods leverage sMRI and fMRI data to construct brain graphs that capture disruptions in cortico-subcortical and heteromodal networks implicated in both motor and cognitive symptoms. The reported prevalence of PDD varies widely, ranging from 24\% to 90\%, depending on disease duration, cohort characteristics, and diagnostic criteria~\cite{bib90p, bib158, Aarsland2010}. Moreover, PD shares pathophysiological features with LBD, particularly disruptions in the cerebral cortex, motivating integrative approaches across related disorders~\cite{bib93p}. In clinical practice, PDD is frequently associated with neuropsychiatric symptoms such as hallucinations, cognitive fluctuations, and sleep or mood disturbances~\cite{bib94p}, which may further correlate with disruptions in specific brain regions detectable via neuroimaging. Consequently, multimodal datasets incorporating imaging, clinical scores, demographic profiles, and in some cases genetic information, have been employed to improve classification accuracy and subtype differentiation. Recent GNN research has begun integrating XAI techniques to enhance interpretability, enabling the identification of salient brain regions and network features contributing to model predictions. Table~\ref{tab:PD_table} summarises studies that employ these advanced computational approaches for PD diagnosis and progression prediction.

\begin{table}
\centering
    \caption{Research work related to PD diagnosis and prediction.}
    \scriptsize
    \begin{tabular}{p{1.5cm}p{2.0cm}p{2.0cm}p{1.0cm}p{2.5cm}p{2.0cm}p{2.3cm}}
    \toprule
    \textbf{XAI Method} & \textbf{Architecture} & \textbf{Modality} & \textbf{Level} & \textbf{Nature of Explanation} & \textbf{Performance} & \textbf{Application} \\
    \midrule
    \multirow{3}{*}{\parbox{1.5cm}{\raggedright Post-Hoc Explainable}}
    & GNN-Saliency & MRI, DTI & Sub & Gl/In/Sp/Vis+Score & ACC: 95.5\% & Early PD diag.~\cite{bib61bf} \\
    & GAT-Saliency & fMRI & Both & Both/Ex/Sp/Vis+Score & AUC: 0.83 & Symptom clsfn. \& gait severity~\cite{bib80bio} \\
    & GCN-Saliency & EEG & Sub & Gl/In/Sp/Vis+Score & ACC: 90.2\% & Voice-related task~\cite{bib58bf} \\
    \midrule
    \multirow{2}{*}{\parbox{1.5cm}{\raggedright Self Explainable}} 
    & GNN-Built-In & DTI & Sub & Gl/In/Sp/Vis+Score & ACC: 79.6\% & Biomarker \newline discovery~\cite{bib74bf} \\
    & GCN-Built-In & DTI, MRI & Sub & Gl/In/Sp/Score & AUC: 0.95 & Feature based dynamic graph learning~\cite{bib35s} \\
    \midrule
    \multirow{3}{*}{\parbox{1.5cm}{\raggedright Hybrid Explanation}}
    & GAT-Attention \newline \& Saliency & MRI, fMRI, DWI & Sub & Both/In/Sp/Vis+Score & ACC: 86\%  & SC-FC fusion~\cite{bib51dp}\\
    & GNN-\newline GNNExplainer \& \newline Attention & MRI, PET &
    Sub & Both/Both/Ag/Vis+Score & 
    Prec.: 93.4\%, 96.6\%, \newline 85.4\% (MSA, IPD, \newline PSP) & Differential diag. \& \newline identify \newline biomarkers~\cite{ling2025explainable}\\
    & GCN-Causality & QSM & Both & Both/In/Sp/Vis+Score & ACC: 72.93\% & Postural \newline anomalies \newline analysis~\cite{bib46dd} \\
    \bottomrule
    \end{tabular}
    \caption*{\scriptsize 
    \textbf{Level:} Sub = Subject, Popln = Population. \textbf{Nature of Explanation:} Scope/Method/Domain/Mode. Scope: Gl = Global, In = Intrinsic, Sp = Specific, Ag = Agnostic, Vis = Visual. \textbf{Performance:} Prec.= Precision. IPD=Idiopathic Parkinson’s disease, MSA=Multiple system atrophy, PSP=progressive supranuclear palsy.
    }
    \label{tab:PD_table}
\end{table}

\subsection{Post-Hoc Explanation}
In the context of post-hoc explanation, several studies have employed saliency-based approaches to visualize SC-FC regions in the diagnosis of PD. Nerrise et al.~\cite{bib80bio} proposed an explainable geometric-weighted GAT operating on SPD matrices within a Riemannian manifold, achieving an F1-score of 76\% and an AUC of 0.83 for gait severity diagnosis. Their model highlighted the involvement of the sensorimotor, salience, and visual networks. Huang et al.~\cite{bib61bf} developed a multi-task graph structure learning framework for early PD identification, reporting 95\% accuracy. Saliency maps revealed key regions including the hippocampus, precentral gyrus, supplementary motor area, and insula. Zhao et al.~\cite{bib58bf} applied GNNs to voice-related EEG data for PD diagnosis, attaining 90.2\% accuracy and identifying the superior temporal gyrus and Broca’s area as critical regions associated with speech deficits in PD patients.

\subsection{Self Explanation}
In the context of PD, Cui et al.~\cite{bib74bf} introduced IBGNN, an interpretable GNN equipped with a post-hoc explanation generator. The model achieved 79.55\% accuracy on the PPMI dataset, outperforming conventional ML and DL approaches by identifying critical ROIs and connectivity patterns. Li et al.~\cite{bib35s} proposed iGLCN, an interpretable graph learning model for PD diagnosis using multimodal MRI, achieving 91.6\% accuracy and an AUC of 0.950 on the PPMI and Shenzhen datasets. Their model highlighted key regions such as the caudate nucleus and thalamus.

\subsection{Hybrid Explainable}
To diagnose PD, recent studies have adopted hybrid explainability techniques that combine intrinsic and extrinsic approaches, thereby offering more comprehensive interpretability. Safai et al.~\cite{bib51dp} proposed a framework integrating sMRI, DTI, and fMRI to construct multimodal brain connectivity networks, achieving 86\% cross-validation accuracy and 73\% test accuracy. Saliency analysis and attention maps identified key brain regions such as the motor network, basal ganglia, and cerebello-thalamo-cortical network as critical for diagnosis. Ling et al.~\cite{ling2025explainable} introduced an explainable GNN framework that incorporated a transformer-based attention mechanism and Regional Radiomics Similarity Network (R2SN) to construct subject-level metabolic graphs. GNNExplainer was further employed to provide interpretable insights into relevant metabolic regions and connectivity patterns.

Causality has also emerged as a promising approach to guide clinical decision-making, particularly in PD cases with complex motor and non-motor symptoms. Tang et al.~\cite{bib46dd} proposed a causality-driven GCN for diagnosing postural anomalies in PD using QSM imaging, where brain patches were modelled as graph nodes with edges defined by spatial proximity and texture similarity. By leveraging multi-instance learning and causal prediction invariance, their model achieved 72.93\% accuracy and 76.43\% AUC, outperforming traditional radiomic and deep learning methods, while identifying key regions such as the hippocampus and middle frontal gyrus.

\subsection{Discussion}
In PD, saliency-based methods remain a dominant area of research due to their ease of integration and ability to provide visual explanations. For biomarker identification and understanding dynamic feature learning, S-XAI methods demonstrate significant promise, often balancing predictive performance with interpretability. Hybrid approaches, which combine multiple methods and incorporate causal inference, offer a more holistic understanding of disease progression, enabling more nuanced clinical insights.

\section{Multi-Disease Diagnosis}
\label{sec:section8}
Recent studies have demonstrated the potential of XGNNs in distinguishing between different aetiologies of dementia. The ability to discriminate among subtypes such as AD, PD, and FTD is critical, as it enables accurate diagnosis and informed treatment planning. By integrating multimodal data (e.g., MRI, fMRI, EEG), recent work has evaluated multiple dementia subtypes and elucidated their distinct brain network patterns. Such multi-disease analyses contribute to the development of more robust and interpretable GNN models that facilitate differential diagnosis, while also providing insights into the underlying pathophysiology of each condition. By capturing these distinctions, XGNNs enhance diagnostic accuracy and support neurobiological interpretation of dementia subtypes. An overview of recent multi-disease XGNN studies is presented in Table~\ref{tab:Multi_table}.

\begin{table}
\centering
    \caption{Research work related to Multi-disease Diagnosis and Prediction.}
    \scriptsize
    \begin{tabular}{p{1.5cm}p{2.0cm}p{1.0cm}p{1.0cm}p{2.5cm}p{2.3cm}p{2.3cm}}
    \toprule
    \textbf{XAI Method} & \textbf{Architecture} & \textbf{Modality} & \textbf{Level} & \textbf{Nature of Explanation} & \textbf{Performance} & \textbf{Application} \\
    \midrule
    \multicolumn{7}{c}{\textbf{AD vs. FTD}} \\
    \midrule
    \multirow{2}{*}{\parbox{1.5cm}{\raggedright Self Explainable}} 
    & GCN-Built-In & MRI & Both & Both/In/Sp/Vis+Score & ACC: 89.7\% & Differential diag. of AD, FTD, CN~\cite{bib47dd} \\
    & GNN-Attention & fMRI & Sub & Both/Both/Sp/Vis+Score &
    FTD: 87.22\%, \newline ADNI: 89.78\% & Dementia detection \newline and diagnosis~\cite{bib53dp} \\
    \midrule
    \multicolumn{7}{c}{\textbf{AD vs. PD}} \\
    \midrule
    \multirow{3}{*}{\parbox{1.5cm}{\raggedright Self Explainable}} 
    & GNN-Built-In & EEG & Both & Both/In/Sp/Vis+Score & ACC: 97.4\% & AD/PD vs HC diag.~\cite{bib44dd} \\
    & GNN-Attention & fMRI & Both & Both/In/Sp/Vis+Score &TaoWu: 77.5±17.5, \newline PPMI: 64.0±6.63, Neurocon: 68.3 ± 20.0, ADNI: 63.7±2.63 & PD/AD brain network clsfn~\cite{bib65bf} \\
    & GCN-Built-In & fMRI, DTI & Both & Both/In/Sp/Vis& ADNI: ACC 82.3\%; \newline PPMI: ACC 86.2\% & Multimodal brain disease localization~\cite{le2025brainmap}\\
    \midrule
    \multicolumn{7}{c}{\textbf{Amnestic MCI vs. PD}} \\
    \midrule
    \multirow{2}{*}{\parbox{1.5cm}{\raggedright Self Explainable}} 
    & GCN-Attention & MRI & Both & Both/Both/Sp/Vis+Score & HC vs MCI: 84.8±1.92, HC vs PD: 78.3±3.19 & MCI, PD clsfn~\cite{bib45dd} \\
    & GCN-Feature Level & fMRI, DTI 
    & Both & Both/Both/Sp/Vis+Score & MCI vs HC: 90.4±2.4, \newline PD vs HC: 85.9±4.5 & Brain networks for \newline MCI, PD~\cite{bib154} \\
    \bottomrule
    \end{tabular}
    \caption*{\scriptsize 
    \textbf{Level:} Sub = Subject, Popln = Population. \textbf{Nature of Explanation:} Scope / Method / Domain / Mode. Scope: Lo = Local, Gl = Global, Both; Method: In = Intrinsic, Ex = Extrinsic; Domain: Sp = Model Specific, Ag = Agnostic; Mode: Vis = Visual, Score = Score-based, Exmpl = Example-based.
    }
    \label{tab:Multi_table}
\end{table}

\subsection{Evaluating AD and FTD}
In efforts to distinguish AD from FTD, studies have employed neuroimaging-based GNN frameworks to detect disease-specific patterns. Nguyen et al.~\cite{bib47dd} proposed a deep grading framework that combines GCN with sMRI for both disease detection and differential diagnosis. Their model achieved 90.5\% accuracy in differentiating dementia patients from HC, as well as in multi-class classification of AD vs. FTD vs. HC. The learned grading maps highlighted hippocampal abnormalities as salient in AD and ventromedial frontal cortex atrophy in FTD, consistent with known hallmarks of each disorder. These localized atrophy patterns provided clinicians with visual explanations of the model’s decisions. Similarly, Wang et al.~\cite{bib53dp} developed a GNN incorporating self-attention and feature selection to analyze global brain-region interactions from MRI data. Their model identified significant atrophy in the amygdala, precentral gyrus, and parahippocampal gyrus in both AD and FTD compared to HC, underscoring these regions’ relevance in distinguishing the two dementias. The attention weights further served as interpretive cues, highlighting brain regions most influential to the model's predictions. Together, these approaches underscore the utility of brain network-based methods for differential diagnosis. By building on these insights, XGNN models not only achieve high accuracy in classifying AD and FTD but also reveal disorder-specific network disruptions, offering clinically valuable interpretability.

\subsection{Evaluating AD and PD}
Distinguishing AD from PD is another important multi-disease application, as PD patients can develop cognitive impairments that may be mistaken for AD. Cao et al.~\cite{bib44dd} addressed this challenge using a directed structure-learning GNN (DSL-GNN) to classify AD and PD based on EEG-derived effective brain connectivity (EBC) patterns. By integrating univariate features (power spectral densities) with multivariate EBC features, their model leveraged both local and network-level EEG biomarkers. The DSL-GNN achieved strong classification performance: 94.0\% for AD vs. HC, 94.2\% for PD vs. HC, and 97.4\% for direct AD vs. PD classification; it also reached 93.0\% in three-way AD vs. PD vs. HC discrimination. Incorporating directed connectivity, which reflects causal interactions in EEG networks, improved performance over conventional undirected GNNs. Visualisation of learned features revealed distinct disease-specific patterns: AD patients showed pronounced parietal disruptions, while PD patients exhibited temporal and frontal abnormalities, consistent with posterior cortical atrophy in AD and frontostriatal dysfunction in PD.

Xu et al.~\cite{bib65bf} introduced ContrastPool, a contrastive graph pooling method for classifying brain networks from rs-fMRI data across AD and PD cohorts. To address challenges such as low signal-to-noise ratios and small sample sizes, the model applied dual attention mechanisms at both ROI and subject levels. By contrasting patient and control groups during pooling, ContrastPool highlighted the most discriminative regions. It outperformed 21 state-of-the-art GNN and machine learning models, with accuracies of 67.8\% on ADNI (AD vs. MCI vs. CN), 64.0\% on PPMI (early PD vs. controls), and 77.5\% and 75.0\% on the Taowu and Neurocon PD datasets, respectively. Although these accuracies were lower—reflecting the difficulty of multi-class and multi-cohort fMRI classification—the model’s interpretability was a key advantage. The posterior cingulate and precuneus were highlighted as critical in AD, while the temporal cortex was emphasised in PD, aligning with clinical evidence. Le et al.~\cite{le2025brainmap} proposed BrainMAP, a multimodal GCN framework that integrates fMRI and DTI through atlas-guided subgraph filtering and attention-gated fusion. The model achieved strong performance on ADNI (82.3\%) and PPMI (86.2\%)  while maintaining low computational cost. Also, BrainMAP localized disease-relevant ROIs such as subcortical and limbic regions, providing clinically meaningful explainability in AD and PD.

Collectively, these studies demonstrate that XGNNs can detect subtle connectivity differences in noisy EEG and fMRI data and present them in an interpretable manner, thereby supporting differential diagnosis between a primarily cognitive disorder (AD) and a movement disorder (PD) with overlapping dementia-related symptoms.

\subsection{Evaluating Amnestic MCI and PD}
Beyond classic dementia diagnoses, XGNNs have also been applied to distinguish amnestic MCI (aMCI), often considered a prodromal stage of AD, from PD. This comparison is clinically valuable, as both MCI and early PD can present with mild cognitive changes. Yang et al.~\cite{bib154} proposed a Multimodal Dynamic GCN (MDGCN) to jointly capture SC and FC patterns from MRI and DTI data. The model employs a bilateral GCN architecture with a correspondence matrix to dynamically fuse intermodal information, linking structural and functional brain networks at the individual level. This approach accounts for disorder-specific differences in SC–FC coupling and outperformed static and unimodal baselines in classifying MCI, PD, and healthy controls. Attention-weight analysis further revealed distinct network involvement: the right posterior cingulate gyrus within the DMN emerged as a key discriminator for MCI, whereas the left somatomotor cortex was most indicative of PD. By dynamically integrating multimodal data, the MDGCN provided interpretable mappings of divergent structure–function relationships across disorders.

Building on multimodal fusion, Guo et al.~\cite{bib45dd} introduced a Graph-Based Fusion (GBF) framework that integrates imaging, genetic, and clinical data. The model incorporates an imaging–genetic fusion module with attention mechanisms and a multi-graph GCN for final classification. Applied to a dataset of MCI, PD, and controls, the GBF achieved 84.8\% accuracy for MCI and 78.3\% for PD, outperforming single-modality baselines. Attention-weight visualisation yielded biologically meaningful insights: top brain regions for MCI included the rostral middle frontal gyrus, superior frontal gyrus, inferior parietal lobule, and fusiform gyrus—areas implicated in early AD-related decline. For PD, key regions included the rostral anterior cingulate, superior temporal gyrus, superior frontal gyrus, and inferior parietal lobule. Notably, the inferior parietal lobule was salient in both groups, suggesting it may serve as a shared locus of neurodegeneration in early cognitive impairment. These findings reinforce the validity of the model’s predictions by highlighting regions consistent with known MCI- and PD-related pathology.

\subsection{Discussion}
The reviewed studies underscore that attention-based GNN architectures are particularly well suited for classifying neurodegenerative disorders, owing to their ability to learn node-level and region-specific importance. A consistent trend is the benefit of modality fusion in multi-disease models. Integrating data from functional and structural imaging, electrophysiology, genetics, and clinical measures often yields more robust classifiers than those based on a single modality. The success of Guo et al.’s~\cite{bib45dd} imaging–genetic–clinical fusion and Yang et al.’s~\cite{bib154} SC–FC dual GCN demonstrates that cross-domain brain network integration enhances model robustness and captures complementary disease signatures.

At the same time, these complex architectures highlight the importance of effective feature selection and alignment. Techniques such as correspondence matrices to link modalities or attention mechanisms to weight modality contributions are critical for mitigating noise and redundancy. As multi-centre datasets are increasingly leveraged to improve sample size, performance fluctuations across cohorts have become apparent. This underscores the necessity of domain adaptation and data harmonisation in multi-site studies. For instance, a model trained on ADNI MRI data may fail to generalise to another cohort because of differences in scanner characteristics or demographic distributions, which can lead to reduced accuracy. Overall, there is a clear shift toward explainable multimodal graph models of the brain, representing an important step toward actionable AI in clinical practice. An explainable model not only predicts, for example, “this patient likely has FTD,” but also provides a neurobiological rationale, such as “because frontal and anterior temporal network connectivity is severely disrupted, whereas hippocampal connectivity is relatively preserved.” Such interpretability directly supports differential diagnosis and enhances clinical decision-making.

\section{Public Datasets for Dementia Research}
\label{sec:section9}
Among the most widely used large-scale imaging datasets for dementia research are ADNI, OASIS, AIBL, MIRIAD, NIFD, and PPMI, each providing unique strengths for investigating neurodegenerative disorders. ADNI remains a cornerstone in Alzheimer’s research, offering multimodal imaging, including structural and functional MRI together with longitudinal data that track progression from healthy controls to AD. OASIS complements this resource by providing both cross-sectional and longitudinal MRI scans, as well as clinical assessments of older adults with varying degrees of cognitive decline.

Building on these foundations, AIBL and MIRIAD incorporate PET and MRI together with genetic data, fluid biomarkers, and cognitive testing, enabling a more comprehensive investigation of AD-related risk factors. Beyond Alzheimer’s disease, the NIFD dataset focuses on FTD, a less prevalent but clinically important subtype characterized by early-onset neurodegeneration. NIFD includes structural and functional imaging, genetic profiles, and clinical measures to support research into FTD progression. For Parkinson’s disease, the PPMI dataset has become a key benchmark, offering sMRI, DTI, and fluid biomarkers to facilitate understanding of PD pathology.

The key characteristics of these datasets, including imaging modalities, targeted disease subtypes, and representative references, are summarised in Table~\ref{table2}.

\begin{table}
\caption{A summary of publicly available neuroimaging datasets used in dementia research.}
\label{table2}%
\scriptsize
     \begin{tabular}{@{}p{4cm}p{3cm}p{3.3cm}p{3.5cm}@{}}
\toprule
\textbf{Dataset} & \textbf{Modalities} & \textbf{Subtype} & \textbf{References} \\
\midrule
Alzheimer’s Disease Neuroimaging Initiative (ADNI) \cite{jack2008alzheimer} & MRI, fMRI, PET, Clinical Assessments, Genetic Data, Biochemical Data & Alzheimer’s Disease and progression & \cite{bib43s}, \cite{bib37s}, \cite{bib68bf}, \cite{bib21s}, \cite{bib22s}, \cite{bib30s}, \cite{bib24s}, \cite{bib153dp}, \cite{bib77bio}, \cite{bib34s}, \cite{bib50dp}, \cite{bib67bf}, \cite{bib55dp}, \cite{bib52dp}, \cite{bib70bf}, \cite{bib40s}, \cite{bib64bf}, \cite{bib31s}, \cite{bib75bio}, \cite{bib25s}, \cite{bib59bf}, \cite{bib76bio}, \cite{bib26s}, \cite{bib71bf}, \cite{bib72bf}, \cite{bib82bio}, \cite{bib33s}, \cite{bib28s}, \cite{bib78bio}, \cite{bib29s}, \cite{bib57bf}, \cite{chen2025guiding}, \cite{bib79bio}, \cite{bib48dp}, \cite{bib81bio}, \cite{bib41s}, \cite{bib66bf}, \cite{bib152dd}, \cite{bib32s}, \cite{bib11}, \cite{bib47dd}, \cite{bib53dp}, \cite{bib65bf}, \cite{bib45dd}, \cite{bib154}\\
\midrule
Open Access Series of Imaging Studies (OASIS) \cite{lamontagne2019oasis} & Structural MRI & Aging, dementia, and brain structure & \cite{bib68bf}, \cite{bib50dp}, \cite{bib82bio}, \cite{bib28s}, \cite{bib39s}, \cite{bib47dd} \\
\midrule
Australian Imaging, Biomarkers {\&} \newline Lifestyle (AIBL) \cite{ellis2009australian} &
MRI, PET, Cognitive Testing & Alzheimer’s Disease and risk factors & \cite{bib82bio}, \cite{bib28s}, \cite{bib47dd} \\
\midrule
Minimal Interval Resonance Imaging in Alzheimer’s Disease (MIRIAD) \cite{malone2013miriad} & Structural MRI & Alzheimer's Disease progression & \cite{bib82bio}, \cite{bib47dd} \\
\midrule
Alzheimer's Disease Prediction \newline of Longitudinal Evolution Challenge \newline (TADPOLE) \cite{marinescu2019tadpole} & MRI, PET, CSF Biomarkers, \newline Clinical Assessments & Prediction of Alzheimer's Disease progression & \cite{bib24s}, \cite{bib42s} \\
\midrule
UK Biobank \cite{sudlow2015uk} & 
Structural MRI, fMRI,\newline DTI, Clinical & 
Aging, neurodegeneration, \newline genetic risk factors & \cite{bib52dp} \\
\midrule
European Diffusion Tensor \newline Imaging Study in Dementia (EDSD) \cite{brueggen2017european} & Structural MRI, DTI & Cross-sectional \newline multi-center study & \cite{bib28s} \\
\midrule
Simple Language and Cognition Decline project (SILCODE) \cite{li2019sino} & MRI, fMRI, Language and Cognition Tests, Demographic and Clinical Data & Early detection of AD and cognitive impairment based on linguistic features & \cite{bib21s} \\
\midrule
Shenzhen Mental Health Centre Cohort \cite{liu2020study} & MRI, fMRI, Cognitive Assessments, Demographic Data & Mild Cognitive Impairment and Alzheimer’s Disease & \cite{bib35s}\\
\midrule
Taowu Dataset (Chinese Longitudinal Aging Study - CLAS or Taowu Project) \cite{badea2017exploring} & MRI, fMRI, Demographic and Lifestyle Data, Cognitive Assessments & Aging, Alzheimer’s Disease risk prediction, early MCI detection & \cite{bib65bf}\\
\midrule
Parkinson’s Progression Markers Initiative (PPMI) \cite{marek2011parkinson} & MRI, DTI, PET, Biomarkers & Parkinson’s Disease and related dementia & \cite{bib61bf}, \cite{bib69bf}, \cite{bib74bf}, \cite{bib35s}, \cite{bib65bf}, \cite{bib45dd} \\
\midrule
UK Parkinson's Disease Society Brain Bank \cite{clarke2016uk} & EEG, Neuropathology Data & Parkinson’s Disease and Dementia with LDB & \cite{bib58bf} \\
\midrule
UK Parkinson’s Disease Roadshow \newline Dataset & 
Clinical Assessments, Cognitive \& Motor Evaluations and Genetic Data & 
Parkinson’s Disease, cognitive decline, MCI conversion risk & \cite{bib44dd}\\
\midrule
Neuroimaging in \newline Frontotemporal Dementia (NIFD) \cite{knopman2014nifd} & MRI, PET, Genetic Data, Clinical Assessments & Frontotemporal dementia and related disorders & \cite{bib47dd}, \cite{bib53dp} \\
\bottomrule
\end{tabular}
\end{table}

\section{Open Challenges and Outlook}
\label{sec:section10}

Despite notable progress in diagnosing dementia subtypes, several open research questions remain that must be addressed to ensure the effective integration of these technologies into clinical practice. In this section, we outline key remaining challenges and present our perspectives on future directions.

\subsection{Explainability Challenges in GNNs}
Current benchmarks in graph neural networks lack standardised experimental settings and consistent evaluation measures~\cite{dwivedi2023benchmarking}. While explainability plays a crucial role in building trust in AI systems, only one study~\cite{bib34s} has incorporated correctness as an explainability metric, underscoring the need for robust, quantitative measures. Approaches like Factor Graph-based Interpretable Neural Networks~\cite{li2025factor} exemplify self-explainable architectures, highlighting the potential for intrinsic explainability in GNNs. Future research should prioritise the development of standardised explainability protocols to ensure reliability.

Dynamic GNNs also face significant hurdles. Although spatiotemporal approaches such as temporal attention or graph saliency are increasingly applied to fMRI and EEG data to capture disease-relevant patterns~\cite{bib21s,shehzad2025dynamic}, they are computationally intensive and often lack intuitive interpretability for clinicians. A particular gap lies in temporal explainability, specifically identifying when certain brain regions contribute most to predictions. Similarly, multimodal integration and nonlinear interactions remain problematic. While structural and functional connectivity fusion can capture complementary aspects of brain alterations, it remains unclear which modality drives specific predictions. XAI tools such as GNNExplainer and SHAP can highlight important nodes and edges, but their outputs are often too abstract for clinical interpretation, limiting their applicability in real-world decision-making. Addressing these issues will require modality-specific attribution methods and clinically meaningful explanations.

\subsection{Causal Explanations}
GNNs primarily yield correlation-based functional connectivity, which limits interpretability and actionable insights. The inherent characteristics of neuroimaging data, such as the lack of controlled interventions and the presence of complex dynamic interactions, make it particularly difficult to distinguish causality from correlation. Although some studies have begun integrating causal inference with GNNs~\cite{bib153dp,bib46dd,febrinanto2025refined,febrinanto2023entropy}, this area remains underdeveloped and computationally demanding, requiring further validation. Progress in this direction is essential to enable GNNs to provide meaningful causal explanations of brain connectivity.

\subsection{Neuroimaging Complexity and Patient Heterogeneity}
Neuroimaging reveals substantial heterogeneity across dementia subtypes and individual patients. Biomarkers such as hippocampal atrophy or DMN disruptions vary between dementia types and often overlap, complicating decision boundaries in GNNs. This underscores the importance of explainability to ensure that model decisions align with established anatomical and clinical markers. Moreover, diseases such as AD and PD exhibit considerable inter-subject variability, with some patients primarily driven by A$\beta$ pathology while others show tau or vascular abnormalities. Current models often fail to capture this diversity in patient-specific explanations. Although clustering-based methods~\cite{bib79bio} can stratify patients, they frequently lack interpretable associations with underlying neural mechanisms, while approaches such as SEHG~\cite{huang2025sehg} offer self-explainable GNNs that jointly optimise predictions and explanations. Future work should contextualise individual disease trajectories in light of known heterogeneity.

\subsection{Scalability and Performance Issues}
Explaining GNN predictions for large graphs remains computationally intensive due to operations such as subgraph sampling and methods like GNNExplainer. As graph size increases, computational costs escalate significantly, limiting real-time applicability~\cite{peng2025biologically}. In sparse graphs, explanation faithfulness has been shown to decline by 59.9\% on large datasets~\cite{agarwal2023evaluating}. Mitigating bias, addressing sparsity, and improving scalability are therefore essential to develop more robust and generalisable models.

\subsection{Domain-Specific Needs}
Generic XAI techniques often generate subgraphs or feature scores that fail to align with domain knowledge in neuroscience, reducing clinical trust. For instance, distinguishing causal mechanisms from correlations in brain connectivity remains challenging. Domain-specific models and explainers are needed to provide clinically relevant and biologically plausible explanations, thereby ensuring trustworthiness in medical contexts.

\subsection{LLMs in Model-Level Explanations}
Recently, studies have begun exploring the integration of large language models (LLMs) with XGNNs, as LLMs can support model-level explanations and contribute to early detection and diagnosis of dementia~\cite{li2025can}. Although still in its early stages, this line of research highlights the importance of generating clinically valid, interpretable narratives. For example, Lee et al.~\cite{lee2025alzheimer} combined speech transcription with graph-based vision models to enhance dementia diagnosis, Hu et al.~\cite{bib114trade} employed a variational regularised encoder-decoder GNN for AD risk estimation with relation importance mechanisms, and Arriba-Pérez et al.~\cite{de2024explainable} used LLMs to analyse patient dialogue for detecting early cognitive decline. These studies illustrate the promise of combining GNN reasoning with natural language explanations to improve clinical adoption. Overall, addressing these challenges, including standardisation, temporal and multimodal interpretability, causal reasoning, patient heterogeneity, scalability, domain alignment, and the integration of LLMs, will be critical for advancing explainable graph models in dementia research and ensuring their adoption in clinical practice.

\section{Conclusion}
\label{sec:section11}
Explainable graph neural networks (XGNNs) offer a transformative opportunity to bridge the gap between advanced machine learning methodologies and clinical applicability in dementia research. By modeling the brain as a network and integrating multimodal data, XGNNs provide powerful tools for detecting disease-relevant biomarkers, characterizing subtype-specific connectivity patterns, and supporting differential diagnosis. The interpretability enabled by these models helps overcome key barriers to trust and adoption in healthcare, allowing clinicians to better understand how predictions are generated and how they relate to neuropathology. In this review, we presented a taxonomy of XGNN techniques employed in dementia studies, organized according to their explanation parameters, and examined their applications across various dementia subtypes. We also identified major open challenges, including the lack of standardized evaluation metrics, difficulties in temporal and multimodal interpretability, limited progress in causal reasoning, patient heterogeneity, scalability constraints, and the need for domain-specific alignment. Emerging directions such as the integration of large language models (LLMs) further highlight the evolving landscape of explainable graph-based frameworks. Future research should focus on establishing rigorous evaluation protocols to ensure that XGNN-generated explanations are technically robust and ethically sound. By advancing methods that provide causal and clinically meaningful insights, explainable graph models hold the potential to transform dementia diagnosis and prognosis, enabling earlier detection, more accurate subtype differentiation, and ultimately more personalized therapeutic strategies.

\bibliographystyle{ACM-Reference-Format}
\bibliography{sample-base}

\end{document}